\documentclass{article}

\usepackage{microtype}
\usepackage{graphicx}
\usepackage{subcaption}
\usepackage{booktabs} 

\usepackage{hyperref}

\usepackage[accepted]{icml2025}

\usepackage{amsmath}
\usepackage{amssymb}
\usepackage{mathtools}
\usepackage{amsthm}
\def \RR {{\mathbb{R}}}

\newtheorem{remark}{Remark}

\newtheorem{definition}{Definition}

\newtheorem{lemma}{Lemma}

\usepackage[textsize=tiny]{todonotes}

\usepackage{amsmath,amsfonts,bm}

\def\eqref#1{equation~\ref{#1}}

\def\1{\bm{1}}

\DeclareMathAlphabet{\mathsfit}{\encodingdefault}{\sfdefault}{m}{sl}
\SetMathAlphabet{\mathsfit}{bold}{\encodingdefault}{\sfdefault}{bx}{n}

\def \bx {{\bm x}}

\def \bb {{\bm b}}

\def \bu {{\bm u}}
\def \bw {{\bm w}}
\def \bo {{\bm o}}
\def\bao{{\bar{{{\bm o}}}}}

\def \bW {{\mathbf W}}
\def \bA {{\mathbf A}}
\def \bX {{\mathbf X}}

\def \bS {{\mathbf S}}

\def \bU {{\mathbf U}}

\def \bP {{\mathbb P}}

\def \bp {{\mathbf{p}}}

\DeclareMathOperator*{\argmin}{arg\,min}

\newcommand{\bigCI}{\mathrel{\text{\scalebox{1.07}{$\perp\mkern-10mu\perp$}}}}
\usepackage{url}            
\usepackage{booktabs}       
\usepackage{amsfonts}       
\usepackage{nicefrac}       
\usepackage{microtype}      
\usepackage{xcolor}         

\usepackage{adjustbox}
\usepackage{graphicx}
\usepackage{subcaption}

\usepackage{amsmath}
\usepackage{amssymb}
\usepackage{mathtools}
\usepackage{amsthm}
\usepackage{multirow}
\usepackage{tabularx}
\usepackage{enumitem}
\newcolumntype{Y}{>{\centering\arraybackslash}X}

\usepackage{titletoc}
\usepackage{hyperref}       
\usepackage[capitalize,noabbrev]{cleveref}  

\usepackage[utf8]{inputenc} 
\usepackage[T1]{fontenc}    
\usepackage{url}            
\usepackage{booktabs}       
\usepackage{amsfonts}       
\usepackage{nicefrac}       
\usepackage{microtype}      
\usepackage{xcolor}         

\usepackage[normalem]{ulem}
\newcommand\tsout{\bgroup\markoverwith{\textcolor{red}{\rule[0.5ex]{2pt}{0.8pt}}}\ULon}

\newtheorem{prop}{Proposition}

\usepackage{titlesec}
\titlespacing\section{0pt}{0pt plus 0pt minus 2pt}{0pt plus 0pt minus 2pt}
\titlespacing\subsection{0pt}{0pt plus 0pt minus 2pt}{0pt plus 0pt minus 2pt}
\titlespacing\subsubsection{0pt}{0pt plus 0pt minus 2pt}{0pt plus 0pt minus 2pt}

\AtBeginDocument{%
  \addtolength\abovedisplayskip{-0.2\baselineskip}%
  \addtolength\belowdisplayskip{-0.2\baselineskip}%
 \addtolength\abovedisplayshortskip{-0.2\baselineskip}%
 \addtolength\belowdisplayshortskip{-0.2\baselineskip}%
}

\icmltitlerunning{Improving Routing in Sparse Mixture of Experts with Graph of Tokens}

\begin{document}

\twocolumn[
\icmltitle{Improving Routing in Sparse Mixture of Experts with Graph of Tokens}




\begin{icmlauthorlist}
\icmlauthor{Tam Nguyen}{rice}
\icmlauthor{Ngoc N. Tran}{vandy}
\icmlauthor{Khai Nguyen}{txam}
\icmlauthor{Richard G. Baraniuk}{rice}
\end{icmlauthorlist}

\icmlaffiliation{rice}{Department of Electrical \& Computer Engineering, Rice University, Texas, USA}
\icmlaffiliation{vandy}{Department of Computer Science, Vanderbilt University, Tennessee, USA}
\icmlaffiliation{txam}{Department of Statistics and Data Sciences, University of Texas at Austin, Texas, USA}

\icmlcorrespondingauthor{Tam Nguyen}{mn72@rice.edu}

\icmlkeywords{Machine Learning, ICML}

\vskip 0.3in
]



\printAffiliationsAndNotice{}  

\begin{abstract}
Sparse Mixture of Experts (SMoE) has emerged as a key to achieving unprecedented scalability in deep learning. By activating only a small subset of parameters per sample, SMoE achieves an exponential increase in parameter counts while maintaining a constant computational overhead. However, SMoE models are susceptible to routing fluctuations--changes in the routing of a given input to its target expert--at the late stage of model training, leading to model non-robustness. In this work, we unveil the limitation of SMoE through the perspective of the probabilistic graphical model (PGM). Through this PGM framework, we highlight the independence in the expert-selection of tokens,
which exposes the model to routing fluctuation and non-robustness. Alleviating this independence, we propose the novel Similarity-Aware (S)MoE, which considers interactions between tokens during expert selection. We then derive a new PGM underlying an (S)MoE-Attention block, going beyond just a single (S)MoE layer. Leveraging the token similarities captured by the attention matrix, we propose the innovative Attention-Aware (S)MoE, which employs the attention matrix to guide the routing of tokens to appropriate experts in (S)MoE. We theoretically prove that Similarity/Attention-Aware routing help reduce the entropy of expert selection, resulting in more stable token routing mechanisms. We empirically validate our models on various tasks and domains, showing significant improvements in reducing routing fluctuations, enhancing accuracy, and increasing model robustness over the baseline MoE-Transformer with token routing via softmax gating.

\end{abstract}
\section{Introduction}
\label{sec:intro}
Mixture of Experts (MoE)~\citep{jacobs1991adaptive,jordan1994hierarchical} has been widely used to scale up the number of parameters of deep neural networks while maintaining efficient computational overhead. MoE appears in applications of deep learning including language processing~\citep{devlin2018bert,radford2019language,raffel2020exploring,kaplan2020scaling,brown2020language,touvron2023open}, vision understanding~\citep{neil2020transformers,bao2021beit,bao2022vlmo,li2023blip,bai2024sequential}, speech processing~\citep{gaur2021mixture}, and other applications~\citep{subramanian2024towards,gormley2011mixture}. 
A recent variation of the Mixture of Experts (MoE) called Sparse MoE (SMoE)~\citep{shazeer2017outrageously}, has been introduced to enhance model size to billion-parameter while maintaining constant computational costs by modularizing the network and activating only specific subsets of experts for each input. SMoE has been applied successfully across various machine learning tasks. For model training, it has been used in pre-training~\citep{fedus2022switch,artetxe2021efficient} and fine-tuning tasks. In terms of applications, it has shown effectiveness in machine translation~\citep{lepikhin2020gshard}, image classification~\citep{riquelme2021scaling}, and large language modeling~\citep{du2022glam}.

\subsection{Sparse Mixture of Experts}
\label{subsec:SMoE}

Mixture of Experts (MoE) introduces dynamic routing into deep learning models by replacing components such as feed-forward or convolutional layers with a set of specialized neural networks called experts. While this approach enhances model capacity and flexibility, it comes with a significant computational overhead, as the model needs to maintain and process multiple expert networks. Sparse Mixture of Experts (SMoE) addresses this limitation by activating only a subset of experts for each input, substantially reducing computational costs while maintaining performance.

Let $E$ be the number of experts in MoE. For each input token \( \bu_i \in \mathbb{R}^D \) of MoE, a router's component $e$-th computes the affinity scores between \( \bu_i \) and expert $e$-th as $\gamma_e(\bu_i)$ and each corresponding expert network $\bm{g}_e$ processes $\bu_i$ to obtain the output $\bm{g}_e(\bu_i)$, where  $e \in [1, 2, \dots, E]$. In practice, the router $\bm{r}(.)$ is often chosen as \( \bm{r}(\bu_i) = [r_{1}(\bu_i),\dots,r_{E}(\bu_i)] = \mathrm{softmax}([\gamma_1(\bu_i), \dots, \gamma_K(\bu_i)]^\top) = \mathrm{softmax}(\bW\bu_i + \bb) \), where \( \bW = [\bw_1,\dots,\bw_E]^{\top} \in \mathbb{R}^{K \times D} \), \( \bb = [b_1,\dots, b_E]^{\top} \in \mathbb{R}^K \), and $\gamma_e(\bu_i) = \bw_{e}^{\top}\bu_i + b_{e}$. We also refer to $\bm{r}(\bu_i)$ as expert scores for token $\bu_i$. MoE aggregates the outputs of all experts as:
\begin{align}
\label{eq:moe}
    \mathrm{MoE}(\bu) = \sum_{e}^E r_e(\bu_i) \bm{g}_e(\bu_i).
\end{align}
To improve computational efficiency, SMoE applies the $\mathrm{TopK}^0$ function, which returns an expert score $r_e(\bu_i)$ only if it ranks among the top-k highest scores and $0$ otherwise. The selected expert scores are then renormalized to sum to $1$. For brevity, we denote the combined $\mathrm{Renormalize}(\mathrm{TopK^0})$ operation simply as $\mathrm{TopK}$ unless otherwise specified.

Outputs from selected experts are then linearly combined:
\begin{align}
    \label{eq:smoe}
    \mathrm{SMoE}(\bu_i) &= \sum_{e}^E \mathrm{TopK}(r_e(\bu_i))\bm{g}_e(\bu_i).
\end{align}
\begin{remark}The routing scores in SMoE can be computed either as $\mathrm{Renormalize}(\mathrm{TopK^0}(r_e(\bu_i)))$ or $\mathrm{Softmax}(\mathrm{TopK}^{\infty}(\gamma_e(\bu_i)))$, where $\mathrm{TopK}^{\infty}$ assigns $-\infty$ to elements outside the $K$ highest affinity scores. These formulations are mathematically equivalent, as proven in Appendix~\ref{subsec:renorm}.
\end{remark}
\textbf{Routing fluctuation in SMoE.}
One of the major challenges in training SMoE is the fluctuating routing decisions during training \citep{dai2022stablemoe,zoph2022st,chi2022representation}. For instance, even in the final epochs of training, up to 33\% of tokens still switch their assigned experts (c.f. Figure~\ref{fig:ent_fluc}). This routing instability leads model non-robustness as small input perturbations can significantly change expert routing decisions, causing different experts to process similar inputs and leading to inconsistent outputs. In addition, improving the consistency of expert routing decisions is necessary for model training since routing fluctuations, especially in the later stages of training, make it challenging to determine an appropriate stopping point for training. 
Therefore, reinforcing consistent routing decisions enhances model robustness and improves overall performance. Thus,  reinforcing consistent routing decisions enhances the robustness and overall performance of the model.
\subsection{Contributions}
We utilize the attention mechanism and similarity between tokens to reduce the routing fluctuation in SMoE. In particular, from probabilistic graphical model (PGM) perspective underlying the (S)MoE, we show that the independence between the chosen experts and each individual token leads to routing fluctuations. We then propose the novel Similarity-Aware (S)MoE, which promotes the assignment of similar tokens to the same expert. Consequently, Similarity-Aware (S)MoE enables tokens to influence each other’s routing decisions, reducing routing fluctuations, and improving the model’s robustness. We extend the PGM framework to an MoE-Attention block in MoE-Transformer models and propose Attention-Aware, which leverages these dependencies captured in the attention matrix to guide the router's decisions. Our contributions are four-fold:
\vspace{-1em}
\begin{enumerate}[leftmargin=0.4cm]
\item Under the probabilistic graphical model (PGM) perspective, we show that the independence between the chosen experts and each individual token results in routing fluctuations in (S)MoE. 

\item We propose the novel Similarity-Aware (S)MoE that break the independence between the chosen experts and tokens, by building a graph of tokens, to mitigate routing fluctuations.

\item Beyond one layer (S)MoE, we develop the PGM framework to (S)MoE-Attention block in the (S)MoE-Transformer. We proposed Attention-Aware (S)MoE, an extension of Similarity-Aware (S)MoE, which leverage the dependencies between tokens captured by the attention matrix to guide the router's decision. 

\item We theoretically prove that the Similarity-Aware (S)MoE, as well as the Attention-Aware (S)MoE, reduce the entropy in the decision-making process of indecisive tokens. This entropy reduction fosters more confident and consistent expert assignments.
\end{enumerate}
We empirically corroborate the advantages of Similarity-Aware (S)MoE and Attention-Aware (S)MoE models across various tasks and domains.

\textbf{Organization.} In Section~\ref{sec:method_1}, we begin by presenting the Probabilistic Graphical Model (PGM) of (S)MoE and analyzing its limitations. Based on these insights, we introduce our Similarity-Aware (S)MoE framework in the same section. We then extend our analysis beyond single-layer (S)MoE by examining (S)MoE-Attention blocks within (S)MoE-Transformers, leading to our proposed Attention-Aware (S)MoE in Section~\ref{sec:method_2}. Section~\ref{subsec:entropy_analysis} provides a detailed entropy analysis of these models, followed by comprehensive experimental results across various tasks in Section~\ref{sec:experiment}. We conclude our work in Section~\ref{sec:conclusion}, with supplementary materials provided in the Appendix.

\textbf{Notaion.} To facilitate the understanding of the content of our work, all used notations are explained in Appendix~\ref{sec:notations}.


\section{Routing in SMoE with Graph of Tokens}
\label{sec:method_1}
Probabilistic graphical models (PGM) provide a framework to understand conditional dependencies of and perform inference on variables. In this section, we examine the PGM that forms the basis of (S)MoE. From this graphical model perspective, we show the underlying assumptions about the independence between variables, highlighting the limitations that can arise from these assumptions.
\subsection{A Probabilistic Graphical Model for (S)MoE}
\label{sec:pgm_smoe}
\begin{figure}
\centering
\includegraphics[width=0.6\linewidth]{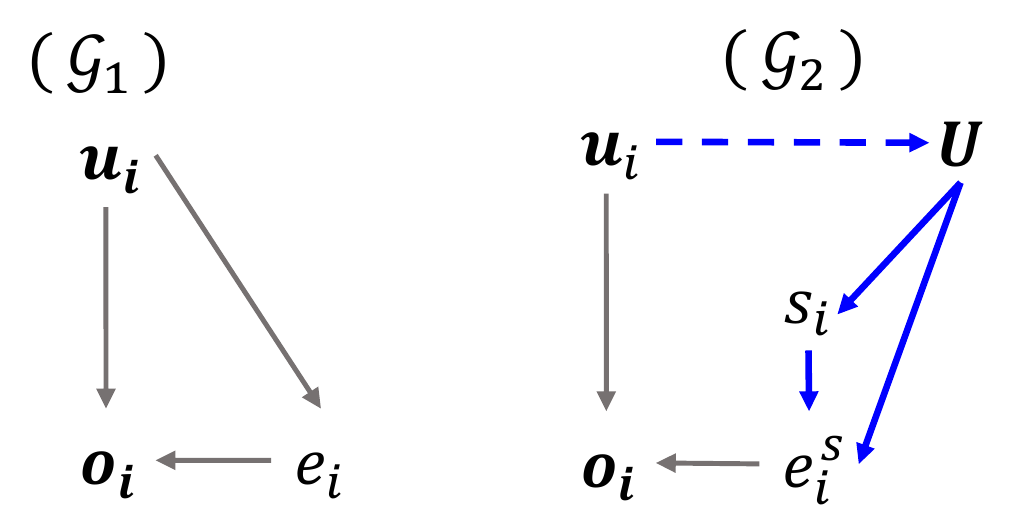}
\vspace{-0.1in}\caption{\small PGMs for (S)MoE ($\mathcal{G}_1$) and Similarity-Aware (S)MoE ($\mathcal{G}_2$). Directed paths are shown by arrows; dotted arrow indicates concatenation; blue arrows highlight differences. $\bU = [\bu_1, \dots, \bu_N]^T$ is the input sequence of (S)MoE and Similarity-Aware (S)MoE. $\bu_i, \bo_i$ are input-output at position $i \in [1,...,N]$. Variables $e_i$ and $e^s_i$ denote expert selection for $\bu_i$ in (S)MoE and Similarity-Aware (S)MoE, respectively; $s_i$ represents similarity variable for $\bU$.}
\label{fig:MIM-illustrate} 
\vspace{-0.2in}
\end{figure}
Given a token $\bu_i$, the expert network $\bm{g}_{e}$. for $e \in [1, \dots, E]$, and the router $\bm{r}(.)$
are defined in Section~\ref{subsec:SMoE}, the generation for (S)MoE~\cite{bishop2003bayesian} takes the form:
\begin{equation}
    \begin{aligned}
        \label{eq:gen_moe}
        e_i|\bu_i &\sim \mathrm{Cat}(\bm{r}(\bu_i)) \qquad \\
\bo_i|\bu_i,{e_i} & \sim \mathcal{N}(\bm{g}_{e_i}(\bu_i), \mathbb{I}) \qquad 
    \end{aligned}
\end{equation}
The generation process, illustrated as the graph $\mathcal{G}_1$ in Fig.~\ref{fig:MIM-illustrate}, can be summarized as such: (S)MoE generative model samples expert selection $e_i$ for token $\bu_i$ from a categorical distribution with probability $\bm{r}(\bu_i)$.
Then, conditioned on the expert selection $e_i$ and token $\bu_i$, output target $\bo_i$ is generated following a Gaussian distribution with mean $\bm{g}_{e_i}(\bu_i)$.

\textbf{(S)MoE as Optimal Estimation}. When dealing with continuous-valued target variables, regression analysis provides a natural framework for prediction tasks. In such problems, our goal is to predict a target variable given an input variable while  minimizing the expected squared error loss between the true target value and our prediction. Under this squared error criterion, the optimal predictor for such regression tasks is the conditional expectation of the target given the input~\cite{hastie2009elements}. In the case of MoE, the optimal estimate of the output $\bo_i$ is given as:
\begin{equation}
\label{eq:expogivenx}
    \begin{aligned}
        \bao_i &= \mathbb{E}[\bo_i|\bu_i] = \mathbb{E}_{e_i}[\mathbb{E}_{\bo_i}[\bo_i| e_i, \bu_i] | \bu_i \bigl] \\
        &= \sum_{e_i}r_{e_i}(\bu_i) \bm{g}_{e_i}(\bu_i) = \mathrm{MoE}(\bu_i).
    \end{aligned}
\end{equation}
Here, $\mathbb{E}[\bo_i | \bu_i, e_i] = \bm{g}_{e_i}(\bu_i)$. As can be seen in Eqn.~\ref{eq:expogivenx}, the optimal estimation $\bao_i$ of $\bo_i$ matches the formula of MoE in Eqn.~\ref{eq:moe}. SMoE can be derived accordingly by using the $\mathrm{TopK}$ function on the expert scores, resulting in $\mathrm{SMoE}(\bu_i) = \sum_e \mathrm{TopK}(r_e(\bu_i))\bm{g}_e(\bu_i)$. When the expert scores $\bm{r}(\bu_i) = \mathrm{Softmax}(\bW\bu_i + \bb)$, the estimation recover (S)MoE with softmax gating.

\begin{remark}[Deriving Other Routers]
Among the important advantages of the PGM formulation of MoE is the flexibility it offers to derive a variety of routers, including the popular ones such as cosine routing, random routing, ... by modifying the  formula for the parameter $\bm{r}(\bu_i)$ of the Cat distribution in Eqn.~\ref{eq:gen_moe}.
\end{remark}

{\bf Limitations.} From the PGM for MoE, we observe that expert selections for each individual token are mutually independent, i.e., $(e_i\bigCI e_j)$ for all $i, j$, when $\bu_i, \bu_j$ are i.i.d. sampled. This lack of interaction between tokens' decisions can lead to \textit{routing fluctuation}.

In particular, near the end of SMoE training, when the learning rate is significantly small and the model parameters stabilize, we expect minimal changes in routing decisions, given that the approximation function is reasonably smooth and token representations do not change considerably.
However, empirical evidence shows this is not the case. Fig.~\ref{fig:ent_fluc} in Sec.~\ref{sec:experiment} presents an empirical analysis demonstrating that up to \textbf{33\% of tokens still switch their selected experts in the final epoch}, highlighting a persistent instability in routing.
This observation suggests that similar tokens should be routed to the same expert, but the current independent routing does not guarantee this from happening. By letting tokens influencing others' expert selection, we could reduce fluctuation and ensure more stable, consistent routing decisions.

\subsection{Similarity-Aware (S)MoE}
\label{subsec:uinform}
Leveraging token similarity to guide expert selection can both reduce routing fluctuations and facilitate expert learning by presenting less diverse input to each expert. In this section, from the PGM perspective, we introduce the novel Similarity-Aware (S)MoE, where expert decisions are directly dependent based on tokens' similarity. 

\textbf{Probabilistic Graphical Model of Similarity-Aware (S)MoE.} For each token $\bu_i$ in a given sequence $\bU$, we introduce a similarity variable $s_i$, whose probability distrubution quantifies how likely token $\bu_i \in \mathbb{R}^D$ resembles other tokens. This similarity directly influences the expert selection variable, which is denoted as $e^s_i$ for token $\bu_i$. The introduction of this new notation for expert selection aims to distinguish it from the expert selection variable  $e_i$ used in the generative model of (S)MoE. 
The generative process for producing the target $\bo_i \in \mathbb{R}^D$ for each token $\bu_i$ is detailed in Def.~\ref{def:sam}, and illustrated as graph $\mathcal{G}_2$ in Fig.~\ref{fig:MIM-illustrate}. 
\begin{definition}
\label{def:sam}
    [Similarity-Aware MoE Generative Model (SAM).] 
    Given a sequence of token $\mathbf{U} = [\bu_1, \dots, \bu_N]^{\top}$, similarity variable $s_i \in \{1, \dots, N\}$ of token $\bu_i$, expert selection variable of SAM $e^s_i \in \{1, \dots, E\}$, and the generation process in SAM generate target variable $\bo_i$ as follows:
    \begin{equation}
    \vspace{-3mm}
        \begin{aligned}
            \label{eq:gen_sam}
{s}_i|\bU &\sim \mathrm{Cat} \left(\mathrm{Softmax}\Bigl(\frac{\bu_i^\top\bW_s\bU^T}{\tau}\Bigl) \right) \qquad \\
e^s_i|s_i,\bU &\sim \mathrm{Cat}\left(\bm{r}(\bu_{s_i})\right) \qquad \\
\bo_i|\bu_i,e^s_i &\sim \mathcal{N}(\bm{g}_{e^s_i}(\bu_i), \mathbb{I}), \qquad
        \end{aligned}
    \end{equation}
\end{definition}
where $\bu_{s_i}=\bU[s_i,:]$,
$\bW_s \in \mathbb{R}^{D \times D}$ is a learnable parameter matrix, and $\tau > 0$ is a temperature parameter controlling the sharpness of the similarity distribution. In practice, to be computationally efficient, we set $\bW_s = \mathbb{I}$.

\begin{remark}
    We establish a connection between the expert selection variables in $e^s_i$ in SAM and $e_i$ in (S)MoE generative model, as follows:
    \begin{equation}
    \label{eq:sam_routing}
    \begin{aligned}
    &\bP(e^s_i = e| \bU) = \sum_{s_i} \bP(e^s_i = e| s_i, \bU)\bP(s_i| \bU) \\
    &= \sum_{s_i} r_e(\bu_{s_i})\bP(s_i| \bU) =\sum_{s_i} \bP(e_{s_i} = e| \bu_{s_i})\bP(s_i| \bU) .
\end{aligned}
\end{equation}
\end{remark}
Equation~\ref{eq:sam_routing} reveals a key distinction in expert routing mechanisms: \emph{The (S)MoE approach routes token $i$ based solely on its embedding $\bu_i$, while our Similarity-Aware routing considers the relationships between all tokens, weighting decisions by token similarities}. The process implies that similar tokens are more likely to be routed to the same expert, promoting consistency in the processing of related information, leading to reduce routing fluctuation. 



\textbf{Optimal Estimation of $\bo_i$ in SAM.} Similar to the derivation of (S)MoE in Section~\ref{sec:pgm_smoe}, we compute the expectation $\mathbb{E}[\bo_i|\bU]$ as follows:
\begin{equation}
\label{eq:regress_function}
    \begin{aligned}
        \bao_i &= \mathbb{E}[\bo_i|\bU] = \mathbb{E}_{e^s_i}\bigl[\mathbb{E}_{\bo_i}[\bo_i| e^s_i, \bU]| \bU ] \\
        &= \sum_{e = 1}^E\bP(e^s_i = e| \bU)\bigl[\mathbb{E}_{\bo_i}[\bo_i| e^s_i = e, \bu_i]\bigl]\\
        &= \sum_{e = 1}^E\sum_{s_i = 1}^N r_e(\bu_{s_i})\bP(s_i | \bU)\bm{g}_{e}(\bu_i).
    \end{aligned}
\end{equation}
With this result, we now define Similarity-Aware (S)MoE:
\begin{definition}
\label{def:U_SMOE}
[Similarity-Aware (S)MoE.] Given an token sequence $\bU = [\bu_1, \dots, \bu_N]^T$, the expert scores $\bm{r}(\bu_i)$ for each token $i$, and the similarity score $\bS[i, j] = \mathrm{Softmax}\left(\displaystyle\bu_i^T\bW_s\bu_j/\tau\right)$, Similarity-Aware MoE estimates the output $\bo_i$ at token $i$ as
\begin{align}
\nonumber
    \bao_i = \sum_{e = 1}^E\sum_{j = 1}^N\bS[i, j]r_e(\bu_j)\bm{g}_e(\bu_i).
\end{align}
and its special version Similarity-Aware SMoE calculates
\begin{align}
    \bao_i = \sum_{e = 1}^E\mathrm{TopK}\left(\sum_{j = 1}^N \bS[i, j]r_e(\bu_j)\right)\bm{g}_e(\bu_i).
\end{align}
\end{definition}
By incorporating token similarities, encourages experts to specialize in handling clusters of similar tokens, leading to more efficient learning and better performance. In addition, the approach is less likely to make different expert selections for similar tokens, hence, leads to reduction in routing fluctuations and improve robustness.

\section{Beyond One Layer: Routing in MoE Transformer with Graph of Tokens}
\label{sec:method_2}
In this section, we extend the PGM for (S)MoE to a 2-layer block setting, which includes the (S)MoE following a self-attention layer as in recent (S)MoE transformer models, such as Switch Transformer~\cite{fedus2021switch} and Swin-MoE~\cite{liu2021swin}. We then propose another Similarity-Aware (S)MoE, which leverages the attention matrix in self-attention to guide expert selection in (S)MoE.
\subsection{PGM for MoE Transformer}
\label{subsec:mam_pgm}
\subsubsection{Multi-head Attention and MoE} 
{\bf Multihead Attention} For a given input sequence $\bX=[\bx_1,\cdots,\bx_N]^\top\in \RR^{N\times D}$, self-attention transforms $\bX$ into the output sequence 
$\small {\rm Softmax}\Big(\bX\bW_{Q, h}^\top\bW_{K, h} \bX^\top/\sqrt{D}\Big)\bX\bW_{V, h}^\top :=\bA_h\bW_{V, h}$, for each head $h=1,\dots,H$. The matrix $\bA_h \in \RR^{N\times N}$ is called the attention matrix, and $\bW_{Q,h},\bW_{K,h}\in \RR^{D_{qk}\times D}$, and $\bW_{V, h}\in \RR^{D_v\times D}$ are the weight matrices for head $h$. 
 MHA aggregates the output of $H$ heads as
\begin{align}
\label{eq:mha}
    \bU = \mathrm{MHA}(\bX) := \frac{1}{H} \sum_{h= 1}^H\bA_{h} \bX\bW_{V, h}^\top\bW_{O,h},
\end{align}
where $\bW_{O,h} \in \RR^{D_v \times D}$ is the projection matrix for the output of each head $h$. Here, we merge $\bW_h =: \bW_{V, h}^T\bW_{O,h} \in \mathbb{R}^{D\times D}$ for convenience, results in $\mathrm{MHA}(\bX)= \frac{1}{H} \sum_{h}\bA_{h}\bX\bW_h$.

{\bf (S)MoE Transformer.} (S)MoE Transformer integrates (S)MoE into a transformer architecture by replacing the standard feed-forward network following the self-attention layer with an (S)MoE layer. This block, refered to (S)MoE-Attention, computes the output token at position $i$-th as $\mathrm{MoE}(\mathrm{MHA}(\bX)[i])$ or $\mathrm{SMoE}(\mathrm{MHA}(\bX)[i])$, where $\mathrm{MHA}$, $\mathrm{MoE}$, $\mathrm{SMoE}$ are defined in Eq.~\ref{eq:mha}, ~\ref{eq:moe}, and~\ref{eq:smoe}, respectively.




\subsubsection{Graphical model for (S)MoE-Attention}
\begin{figure}[t]
\centering
\vspace{-2mm}
\includegraphics[width=0.7\linewidth]{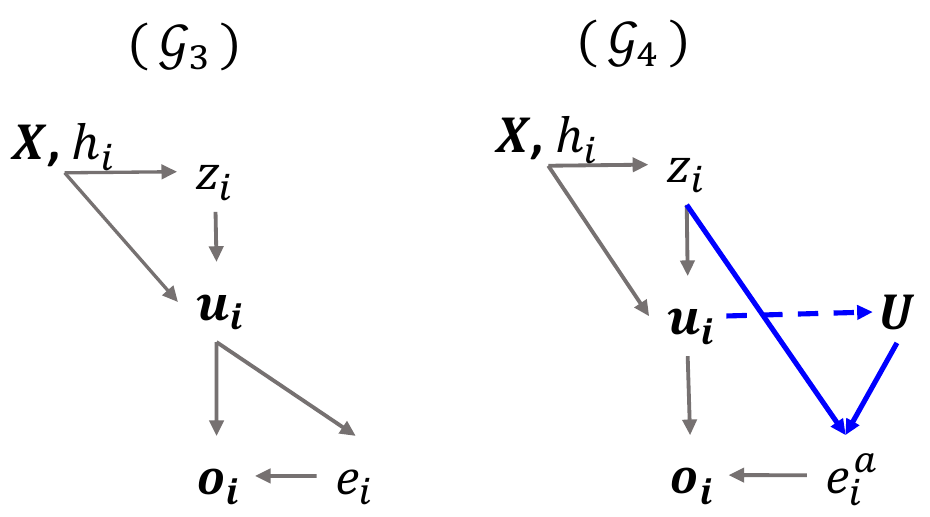}
\vspace{-0.1in}
\caption{\small PGMs for (S)MoE-Attention ($\mathcal{G}_3$) and Attention-Aware (S)MoE ($\mathcal{G}_4$) defined in Def.~\ref{def:mam_pgm} and Def.~\ref{def:aamm_pgm}, respectively. Directed paths shown by arrows; dotted arrow indicates concatenation; blue arrows highlight differences.
}
\label{fig:moe_trans_pgm} 
\vspace{-0.2in}
\end{figure}

Extending the PGM for (S)MoE in Sec.~\ref{sec:pgm_smoe}, we develop the PGM model for (S)MoE-Attention, namely the (S)MoE-Attention Generative Model (MAM), as illustrated as $\mathcal{G}_3$ in Fig.~\ref{fig:moe_trans_pgm}. We define MAM as follows:
\begin{definition}
\label{def:mam_pgm}
    [(S)MoE-Attention Generative Model (MAM).]
    Given the sequence of input $\bX = [\bx_1, \dots, \bx_N]^{\top}$, let $h_i \in \{1, \dots, H \}$ and $z_i \in \{1, \dots, N\}$ be the head selection variable and the attention position selection variable, respectively, for each token $\bx_i$. Let $\bu_i$ be the input token of MoE and $e_i \in \{1, \dots, E\}$ is the expert selection for token $\bu_i$. MAM generates the output $\bo_i$ as follow:
\begin{equation}
\begin{aligned}
\label{eq:def:mam_pgm}
h_i &\sim \mathrm{Uniform}(\{1,\ldots,H\}) \qquad \\
z_i|h_i,\mathbf{X} &\sim \mathrm{Cat}\left(\mathrm{Softmax}\Bigl(\frac{\bx_i^{\top}\bW_{Q, h_i}^{\top}\bW_{K, h_i}\bX^T}{\sqrt{D_{qk}}} \Bigl)\right) \qquad \\
\bu_i|{z}_i, h_i,\bX &\sim \mathcal{N}(\bW_{h_i}\bx_{z_i}, \sigma^2\mathbb{I})\\
        e_i|\bu_i &\sim \mathrm{Cat}(\bm{r}(\bu_i)) \qquad \\
\bo_i|\bu_i,{e_i} & \sim \mathcal{N}(\bm{g}_{e}(\bu_i), \mathbb{I}),
    \end{aligned}
\end{equation}
where $\bW_{Q,h}, \bW_{K,h}, \bW_{h}$ are learnable parameters and $\bx_{z_i} = \bX[z_i,:]$.
\end{definition}
The MAM generation process consists of two main steps: (1) The model generates the input token $\bu_i$, $i=1,\dots,N$ of (S)MoE, from the input sequence $\bX$. In particular, the head selection $h_i$ is uniformly sampled. The attention position $z_i$ is then chosen given the sample $h_i$ and input sequence $\bX$. Subsequently, $\bu_i$ is generated via a Gaussian centered at $\bW_{h_i} \bx_{z_i}$. (2) Following the (S)MoE generation described in Sec.~\ref{sec:pgm_smoe}, for each token $\bu_i$, MAM samples an expert $e_i$ from a categorical distribution with parameter $\bm{r}(\bu_i)$ and generates the output $\bo_i$ from a Gaussian with mean $\bm{g}_{e_i}(\bu_i)$.

\vspace{-0.2in}
\textbf{(S)MoE-Attention as a Point Estimation.} 
We show that MoE-Attention can be derived as an point estimation of the output $\bo_i$ from MAM. In particular, the output $\bo_i$ in MAM can be estimated as 
\begin{equation}
\label{eq:reg_mam}
    \nonumber
    \bao_i = \mathbb{E}[\bo_i | \bX] = \mathbb{E}_{\bu_i}\left[\mathbb{E}_{\bo_i}[\bo_i |  \bu_i]  |  \bX\right] = \mathbb{E}_{\bu_i}\left[\mathrm{MoE}(\bu_i)  |  \bX\right],
\end{equation}
where $\mathbb{E}_{\bo_i}[\bo_i |  \bu_i] = \mathrm{MoE}(\bu_i)$ as in Eqn.~\ref{eq:expogivenx} and $\small \mathbb{E}[\bo_i | \bX] = \mathbb{E}_{\bu_i}\left[\mathbb{E}_{\bo_i}[\bo_i |  \bu_i]  |  \bX\right]$ is obtained by using tower rule. 
Notice that 
$\mathbb{E}_{\bu_i}[\mathrm{MoE}(\bu_i)  |  \bX]$ does not have a closed-form expression. MoE-Attention approximates this conditional expectation using a point estimation of $\bu_i$ given $\bX$, giving:
\begin{align}
\label{eq:reg_moe_a}
\bao_i = \mathbb{E}[\bo_i | \bu_i = \mathbb{E}[\bu_i | \mathbf{X}]] = \mathrm{MoE}(\mathrm{MHA}(\mathbf{X})[i]),
\end{align}
where $\mathrm{MHA}(\bX)$ in Eqn.~\ref{eq:reg_moe_a} is obtained as
\begin{equation}
\label{eq:expugivenx}
\begin{aligned}
&\quad\mathbb{E}[\bu_i | \bX] =\mathbb{E}_{h_i}\bigl[ \mathbb{E}_{z_i}\big[ \mathbb{E}_{\bu_i}[\bu_i |  z_i, h_i, \bX] |  h_i, \bX \bigl]\\
&=\frac{1}{H} \sum_{h=1}^{H} \mathbf{W}_{h} \sum_{j=1}^{N}\bA_h[i, j] \mathbf{x}_{j}= \mathrm{MHA}(\bX),    \nonumber
\end{aligned}
\end{equation}
which is the multihead attention in Eqn.~\ref{eq:mha}.
The detailed derivation of $\mathbb{E}[\bu_i | \bX]$ is given in Appendix~\ref{sec:derivation}.

As can be seen in Eqn.~\ref{eq:reg_mam}, MoE-Attention utilizes MHA to compute the token $\bu_i$ from the input sequence $\bX$. $\bu_i$ is then sent to an MoE layer to estimate the output token $\bo_i$. SMoE-Attention can be easily derived by replacing the $\mathrm{MoE}$ layer in Eqn.~\ref{eq:reg_moe_a} with the $\mathrm{SMoE}$ module in Eqn.~\ref{eq:smoe}, obtaining $\bao_i = \mathrm{SMoE}(\mathrm{MHA}(\bX)[i])$.

{\bf Limitation:} The graphical model $\mathcal{G}_3$ reveals that expert selections for individual tokens exhibit conditional independence given the input tokens, expressed as $(e_i \bigCI e_j | \bX)$ for all $i, j$. This conditional independence and lack of direct interaction in token routing decisions can lead to instability across the network. Our empirical analysis in Fig.~\ref{fig:ent_fluc} (Sec.~\ref{sec:experiment}) demonstrates significant routing volatility, with 10-33\% of tokens changing their assigned experts, across layers, the final training epoch. This volatility arises because conditional independence fails to ensure that similar tokens are routed consistently to the same expert. By introducing inter-token influences in expert selection, we can reduce this fluctuation and achieve more stable, consistent routing patterns.
\subsection{Attention-Aware (S)MoE)}
\label{subsec:attention_informed}
The routing decision for each token can also be informed via their dependency captured in the attention layers. In particular, we establish a link from the variable $z_i$--which represents the position of the token that token $\bx_i$ attends to in the attention layer--to the expert selection variable. This approach, which we call Attention-Aware Routing (A$^{2}$ Routing), allows us to utilize the similarity information directly from the attention layer to inform the expert selection, instead of computing the similarity matrix based on $\bU$ as in Similarity-Aware (S)MoEs.

\textbf{Probabilistic Generative Model for Attention-Routing.}
We define the PGM for Attention-Aware (S)MoE (shown in $\mathcal{G}_4$, Fig.~\ref{fig:moe_trans_pgm}), which employs A$^{2}$ Routing for expert selection: 
\begin{definition}[Attention-Aware MoE Generative Model (A$^{2}$MM)]
\label{def:aamm_pgm}
    Given the sequence of input $\bX = [\bx_1, \dots, \bx_N]^{\top}$, let $h_i \in \{1, \dots, H \}$ and $z_i \in \{1, \dots, N\}$ are the head selection and attention position selection variables of each token at position $i$-th, respectively. Let $\bU = [\bu_1, \dots, \bu_N]^{\top}$ be input tokens to MoE layer and $e^a_i \in \{1, \dots, E\}$ is the expert selection for token $i$. A$^2$MM generates the output $\bo_i$ as follows:
    \begin{equation}
    \begin{aligned}
    \label{eq:moe_mha_pgm}
h_i &\sim \mathrm{Uniform}(\{1,\ldots,H\}) \nonumber \\
z_i|h_i,\mathbf{X} &\sim \mathrm{Cat}\left(\mathrm{Softmax}\Bigl(\frac{\bx_i^{\top}\bW_{Q, h_i}^{\top}\bW_{K, h_i}\bX^T}{\sqrt{D_{qk}}} \Bigl)\right) \nonumber \\
\bu_i|{z}_i, h_i,\bX &\sim \mathcal{N}(\bW_{h_i}\bx_{z_i}, \sigma^2\mathbf{I}) \nonumber\\
e^a_i|z_i,\bU &\sim \mathrm{Cat}(\bm{r}(\bu_{z_i})) \nonumber \\
\bo_i|\bu_i,e^a_i &\sim \mathcal{N}(\bm{g}_{e^a_i}(\bu_i), \mathbf{I}), \nonumber
\end{aligned}
\end{equation}
\end{definition}
where $\bx_{z_i} = \bX[z_i,:]$ and $\bu_{z_i} = \bU[z_i,:]$.
A$^2$MM generates outputs in two steps: (1) Create the sequence of tokens $\mathbf{U}=[\bu_1, \dots, \bu_N]^{\top}$ via multi-head attention, where each $\bu_i$ is generated from the input sequence $\bX$ following the process in Def.~\ref{def:mam_pgm}. (2) For each token $\bu_i$, sample expert $e^a_i$ from a categorical distribution with parameters $\bm{r}(\bu_{z_i})$, where $z_i$ indicates the attended position. The final output $\bo_i$ is drawn from $\mathcal{N}(\bm{g}_{e^a_i}(\bu_i), \mathbb{I})$.


\textbf{Estimation of the Target Values $\bo_i$.} 
By using tower rule, we calculate the conditional expectation $\mathbb{E}[\bo_i|\bX]$ under the A$^2$MM as follows:
\begin{align}
\label{eq:regfa}
    \bao_i = \mathbb{E}[\bo_i|\bX] = \mathbb{E}_{\bU}\left[\mathbb{E}_{e^a_i}\Bigl[\mathbb{E}_{\bo_i}[\bo_i| e^a_i, \bu_i, \bX] | \bU, \bX \Bigl]| \bX\right].
\end{align}
Lemma~\ref{lem:finalda} provides the key result for computing the expectation in Eqn.~\ref{eq:regfa}. The details derivation of Lemma~\ref{lem:finalda} is found in Appendix \ref{sec:prooflem1}
\begin{lemma}
\label{lem:finalda}
    The distribution of the expert selection $e^a_i$ conditioned on $\bU, \bX$, is given by
    \begin{equation}
    \label{eq:lemma1}
        \bP(e^a_i = e | \bU, \bX) = \sum_{h = 1}^H \sum_{j = 1}^N \bold{H}^p[i, h]\bold{A}^p_{h}[i,j]r_e(\bu_j), \nonumber
    \end{equation}
where 
the posteriors
{
\begin{align*}
\bold{A}^{p}_{h}[i,j] &:= \bP(z_i = j  | h_i = h, \bu_i, \bX) \\
&= \displaystyle \frac{\bold{A}_{h}[i, j]\mathbf{L}_{h}[i, j]}{\sum_{j'}\bold{A}_{h}[i, j']\mathbf{L}_{h}[i, j']},
\\
\bold{H}^{p}[i, h] &:=\bP(h_i = h  | \bu_i, \bX) \\
&= \displaystyle\frac{\bold{H}[i, h] \sum_{j}\bold{A}_h[i, j]\mathbf{L}_h[i, j]}{\sum{h'}\bold{H}[i, h']\sum_{j'}\bold{A}_{h'}[i, j']\mathbf{L}_{h'}[i, j']},
\end{align*}}
with the prior $\bold{A}_{h}[i, j] := \bP(z_i = j | h_i = h, \bX)$ and $\bold{H}[i, h] := \bP(h_i = h)$ and the likelihood
$\mathbf{L}_{h}[i, j] := \mathcal{N}(\bu_i |\bW_h\bx_{j}, \sigma^2\bold{I})$. 
This results in
\begin{equation}
    \begin{aligned}
\label{eq:reg_aa_lemma1}
    \mathbb{E}[\bo_i|\bX] = \mathbb{E}_{\bU}\left[\sum_{e}\bP(e^a_i = e | \bU, \bX)\bm{g}_{e}(\bu_i)|\bX\right].
\end{aligned}
\vspace{-4mm}
\end{equation}
\end{lemma}
Lemma \ref{lem:finalda} unveils a sophisticated decision-making process in Attention-Aware MoE, where the final \emph{routing decision for a token is influenced by the decisions of other tokens, as well as the relevance of each attention head}. This formulation can be interpreted as a two-stage process: first, each token's original decision is adjusted by the decisions of other tokens, weighted by $\mathbf{A}^p_h[i,j]$, which represents the ``responsibility'' of token $\bx_j$ in explaining token $\bu_i$'s representation within attention head $h$. Then, these weighted decisions from each head are further weightedly combined by $\mathbf{H}^p[i,h]$, which represents the responsibility of head $h$ in explaining token $\bu_i$. This hierarchical weighting scheme allows the model to integrate context from multiple attention patterns.

{\bf Enhancing Efficiency of Estimating $\bo_i$.} Computing the RHS of Eqn~\ref{eq:reg_aa_lemma1} in Lem.~\ref{lem:finalda} is costly due to the summation over the head $h$. To reduce this computational overhead, we propose an approximation for 
$\mathbf{H}^p$ that avoids full posterior inference across all heads:
\begin{align} 
\label{eqn:h_eff_est}
\bar{\bold{H}}^p[i, h] = \begin{cases}
        1,  \text{if } h = h^{*}:= \argmin_{h} \mathbb{E}[\mathcal{H}(\bA_h[i,:])]\\
        0, \hspace{.15cm}\text{otherwise}.
    \end{cases}
\end{align} 
where $\bA_h[i,:]$ is the $i^{th}$ row of $\bA_h$ and $\mathcal{H}(\bA_h[i,:])= -\sum_{i=j}^N \bA_h[i,j] \log \bA_h[i,j]$ is the entropy of attention score for token $\bx_i$ at head $h$ and the expectation $\mathbb{E}[\mathcal{H}(\bA_h[i,:])]$ is taken over tokens $\bx_i$. This means that only the attention head with the lowest average entropy should contribute to the posteriors. 
Finally, since the expectation over $\bU$ in Equation~\ref{eq:reg_aa_lemma1} does not have a closed-form, we approximate it by using the point estimate $\bU = \mathbb{E} [\bU  | \bX] = \mathrm{MHA(\bX)}$ as in derivation . By applying the result in Lem.~\ref{lem:finalda} and the head selection in Eqn.~\ref{eqn:h_eff_est}, the target values $\bo_i$ is then estimated as:
\begin{align}
\label{eq:o-a}
    \bao_i = \sum_{e = 1}^E\sum_{j = 1}^N\bold{A}^p_{h^*}[i,j]r_e(\bu_j)\bm{g}_e(\bu_i).
\end{align}
We are now ready to define Attention-Aware (S)MoE.
\begin{definition}[Attention-Aware (S)MoE.]
    Given a sequence of input tokens $\bX$, the output of the multihead attention layer $\bU = \mathrm{MHA}(\bX) = [\bu_1, \dots, \bu_N]^T$, the expert score $\bm{r}(\bu_i)$, $i = [1, \dots, N]$, computed in Sec.~\ref{subsec:SMoE}, and the posterior score $\bA^{p}_{h^*}$ computed as in Lemma~\ref{lem:finalda} with $h^*$ being the head index with lowest average attention as in Eqn.~\ref{eqn:h_eff_est}. The Attention-Aware SMoE approximates the output $\bo_i$ as
\begin{align}
    \bao_i = \sum_{e = 1}^E \mathrm{TopK}\bigl(\sum_{j = 1}^N\bA^{p}_{h^*}[i, j]{r_e(\bu_j)}\bigl)\bm{g}_e(\bu_i).
\end{align}
\end{definition}

\section{Entropy Analysis of Similarity and Attention-Aware Routing}
\label{subsec:entropy_analysis}
When the model is uncertain in its routing decision, a small perturbation in either weight space or input space would cause a change in its discrete decision. As a result, high entropy of the expert scores $\bm{r}(\bu_i)$, as defined in Section~\ref{subsec:SMoE}, of a token suggests an increased routing fluctuation. In this section, we demonstrate that our Similarity/Attention-Aware MoE reduces routing fluctuations by lowering the entropy of routing scores.

Consider for any $i = 1, \dots, N$, and define $J_i = \{j \mid \mathcal{H}(e_j| \bu_j) \leq \mathcal{H}(e_i| \bu_i)\}$. Here, we slightly abuse the notation of entropy $\mathcal{H}$, using it interchangeably for both a random variable and its associated distribution.
We apply the Similarity/Attention-Aware MoE to token $\bu_i$ with the set $J_i$. The score function $s(i, j)$, capturing the correspondence between token $\bu_i$ and $\bu_j$ for $j \in J_i$, is either defined as $s(i, j) = \mathrm{Softmax}\left(\displaystyle \bu_i^T\bW_s\bu_j/{\tau}\right)$ (in Def.~\ref{def:sam}) or $s(i, j) = 
\bA^p_{h^*}[i, j]$ as in Eqn.~\ref{eq:o-a}.
We show that the expert selection of Similarity/Attention-Aware (S)MoEs have lower entropy than those of MoE:
\begin{prop}
\label{prop:entroy}
Let $\bm{p}_i = [p_1, \dots, p_K]^T$ denote the distribution of expert selection variables, i.e., $e^s_i$ for SAM and $e^a_i$ for A$^2$MM. The expert score in the baseline MoE for token $\bu_i$ is breviated as $\bm{r}_i := \bm{r}(\bu_i)$ as in Section~\ref{subsec:SMoE}. 
Similarity/Attention-Aware MoE transform these expert  scores $\bm{r}_i$ into $\bm{p}_i = \sum_{j \in J_i} s(i,j)\bm{r}_j$, where $s(i,j)$ denotes the influence weight between tokens $\bu_i$ and $\bu_j$. 
The upper bound of the expert selection's entropy in Similarity/Attention-Aware MoE is then given by:
    \begin{equation}
        \mathcal{H}(\bp_i) \leq \sum_{j = 1}^{|J_i|}s(i, j)\mathcal{H}(\mathbf{r}_j) + \mathcal{H}(\bold{s}_i),
    \end{equation}
    where $\bold{s}_i = [s(i, 1), \dots, s(i, |J_i|)]^T$. As the temperature parameter $\tau \to 0$ (defined in Def.~\ref{def:sam}) or $\sigma \to 0$ (defined in Def.~\ref{def:aamm_pgm}), $\mathcal{H}(\bm{p}_i) \leq \mathcal{H}(\bm{r}_i)$. 
\end{prop}
Our upper bound $\mathcal{H}(\bm{p}_i) \leq \mathcal{H}(\bm{r}_i)$ in Prop.~\ref{prop:entroy} shows that the entropy of the expert scores reduces when applying our methods. Thus, the model improves its decision certainty, reducing the fluctuation in token routing.
\section{Experimental Results}
\label{sec:experiment}
To highlight the strengths of Similarity/Attention-Aware SMoE, we conduct experiments on ImageNet classification, Wikitext-103 language modeling, and fine-tuning tasks. Our results demonstrate that the proposed models: (1) consistently outperform baseline SMoE across all tasks; (2) achieve significant robustness improvements when evaluated on both adversarially and naturally perturbed versions of these datasets; (3) exhibit adaptivity; and (4) surpass previous works while serving as effective plug-and-play enhancements for various MoE architectures. Additionally, empirical analysis reveals that our methods: (5) reduce expert decision entropy and routing fluctuations compared to the baseline, and (6) enhance load balancing for expert assignment. Additional experiments and ablation studies, which further demonstrate the advantages of our methods, are presented in Appendix~\ref{sec:expdetails}.

\textbf{Language modeling on Wikitext-103.}
Table \ref{tab:segment} demonstrates that our Similarity/Attention-Aware SMoEs outperform SMoE and GLAM baselines on Wikitext-103 LM, using TopK experts (K=2). The models are evaluated on validation and test sets using perplexity scores (lower is better) on clean data and in adversarial scenarios, i.e. under word-swap attacks. Our proposed models consistently outperform both baselines. Especially, GLAM variants with Similarity/Attention-Aware SMoEs also show significant improvements over standard versions, demonstrating the approach's effectiveness for both performance and robustness.
\begin{table}[t!]
\vspace{-4mm}
\centering
\caption{\small PPL evaluation (lower is better) with the clean and attacked Wikitext-103 test set Baseline SMoE, X-MoE, SMoE-dropout and their Similarity/Attention-Aware variants, $K = 2$}
    \vspace{-0.4em}
    \color{black}
    \begin{center}
    \scalebox{.66}{
    \begin{tabular}{@{}lcccc@{}}
    \toprule
         \multirow{2}{*}{Model/Metric} & \multicolumn{2}{c}{Clean Wikitext-103} & \multicolumn{2}{c}{Attacked Wikitext-103} \\
         \cmidrule(r){2-3}\cmidrule(r){4-5}
          & Valid PPL & Test PPL & Valid PPL & Test PPL\\
        \midrule
        \midrule
        \midrule
        \it SMoE  & 33.29  & 34.84  & 41.75  & 43.59 \\
        Similarity-Aware SMoE & \bf 30.75 & \bf 32.03 &  \bf 38.33 & \bf 39.92\\
        Attention-Aware SMoE & 31.31  & 32.23 &  39.68 & 40.91\\
        \midrule
        \midrule
        \it GLAM &  37.55 & 39.10 & 48.01 & 49.75\\
        Similarity-Aware GLAM & 33.72 & 34.92 &  42.19 & 43.72\\
        Attention-Aware GLAM &  35.17  & 36.71  & 44.17  & 45.85\\
        \midrule
        \midrule
        X-MoE &  33.05 &	34.49 &	41.68	& 42.96
        \\
        Similarity-Aware X-MoE & \textbf{31.83}	& \textbf{33.06}&	\textbf{39.92}	&\textbf{41.28}\\ 
        Attention-Aware X-MoE & 32.06 &	33.24 &	40.35 &	41.73\\
        \midrule
        \midrule
        SMoE-dropout &  33.08&	34.67&	41.11&	43.09 
        \\
        Similarity-Aware SMoE-dropout  & 32.47	&\textbf{33.69}&	40.6	&\textbf{41.99}\\
        Attention-Aware  SMoE-dropout  & \textbf{32.21}	&33.91	&\textbf{40.56}	&42.17\\
        \bottomrule
    \end{tabular}}
    \end{center}
\vspace{-0.28in}
\label{tab:segment}
\end{table}

\textbf{Compare and improve previous works.}
We compare our Similarity/Attention-Aware SMoE with X-MoE~\cite{chi2022representation}, which addresses routing fluctuation, and SMoE-dropout~\cite{chen2023sparse}, which improves upon standard SMoE. As shown in Table~\ref{tab:segment}, our models achieve lower PPL scores on both Clean and Attacked Wikitext-103 datasets. Additionally, integrating our methods with these models, creating Similarity/Attention-Aware X-MoE and Similarity/Attention-Aware SMoE-dropout, further improved their performance, demonstrating our approach's effectiveness as a plug-and-play enhancement for various MoE architectures.

\textbf{ImageNet Classification.} 
As shown in Table \ref{tab:im}, our Similarity-Aware and Attention-Aware variants outperform the baseline V-MoE \citep{riquelme2021scaling} on both clean data and robustness tests, including ImageNet-C (corruptions) \citep{hendrycks2019benchmarking}, ImageNet-A (adversarial) \citep{hendrycks2021natural}, ImageNet-R (out-of-distribution) \citep{hendrycks2021many}, and ImageNet-O (OOD detection) \citep{hendrycks2021natural}, further demonstrate the advantages if our methods.
\begin{table}[t]
\vspace{-4mm}
\centering
\caption{\small Test set accuracy of different ImageNet variants on Baseline SMoE, Attention-Aware SMoE, and Similarity-Aware SMoE. All models are trained only on the original ImageNet dataset.}
\vspace{-4mm}
\begin{center}
\scalebox{0.71}{
\begin{tabular}{lccccc}
\toprule
\textbf{Model} & Params & IN-1K  & IN-R & IN-A & IN-C \\
               &                 & Top-1 ↑     & Top-1 ↑     & Top-1 ↑ & Top-1 ↑  \\
\midrule
\textit{V-MoE (baseline)}        & 297M  & 72.71 & 35.42  &  5.27 & 48.72  \\
Similarity-Aware V-MoE       & 297M &  73.21  & 36.58 & 5.60 & 50.45 \\
Attention-Aware V-MoE & 297M & \bf 73.33 &    \bf 36.66   & \bf 6.78 & \bf 50.85  \\
\bottomrule
\end{tabular}
}
\end{center}
\label{tab:im}
\vspace{-4mm}
\end{table}

\textbf{Finetuning on downstream tasks.}
\begin{table}[t!]
\centering
\vspace{-4mm}
\caption{\small \color{black} Top-1 test accuracy on Stanford Sentiment Treebank 5, 2 (SST5, SST2), and Banking-77 (B77)
finetuning task.}

    \color{black}
    \begin{center}
    \scalebox{.9}{
    \begin{tabular}{lccc}
    \toprule
         Model & SST5 & SST2& Banking-77 \\
        \midrule
        \it SMoE & 36.54 & 70.23&83.96\\
        \it Similarity-Aware SMoE & 37.91&71.72&85.19\\
        \it Attention-Aware SMoE & \textbf{38.89}&\textbf{72.41}&\textbf{85.84} \\
        \bottomrule
    \end{tabular}}
    \end{center}
\vspace{-0.16in}
\label{tab:finetuning_appendix}
\end{table}
We evaluate the adaptivity of pretrained SMoE variants through fine-tuning on SST5~\cite{socher-etal-2013-recursive}, SST2~\cite{socher-etal-2013-recursive}, and Banking-77~\cite{Casanueva2020} datasets. Table~\ref{tab:finetuning_appendix} shows that Attention-Aware SMoE achieves the highest accuracy across all datasets, followed by Similarity-Aware SMoE, while baseline SMoE performs worst. These results demonstrate the advantageous adaptivity of our proposed architectures compared to baseline SMoEs.


\begin{figure}[!t]
\centering
\includegraphics[width=0.9\linewidth]{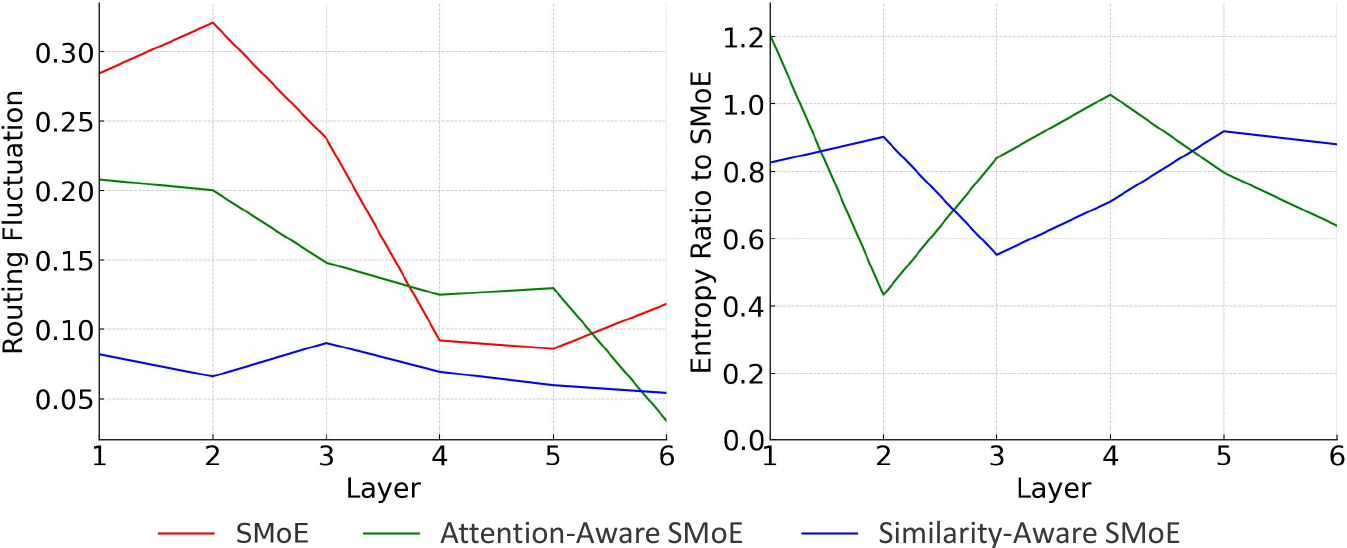}
\vspace{-3mm}
\caption{\small Comparison of routing fluctuation and entropy ratio across layers for Baseline SMoE, Attention-Aware SMoE, and Similarity-Aware SMoE}
\label{fig:ent_fluc} 
\vspace{-.5cm}
\end{figure}
\textbf{Similarity/Attention-Aware MoE reduces routing fluctuation} 
Fig.~\ref{fig:ent_fluc} (Left) compares routing fluctuation across baseline SMoE and our proposed Similarity/Attention-Aware SMoEs on Wikitext-103. Fluctuation rate measures the percentage of tokens changing expert assignments between epochs 59-60. While Baseline SMoE shows highest fluctuation, especially in early layers, Similarity/Attention-Aware SMoEs achieve lower rates throughout. Similarity-Aware SMoE maintains consistently low fluctuation across all layers, demonstrating better routing stability, while Attention-Aware SMoE also significantly improves upon the baseline.

\textbf{Similarity/Attention-Aware MoE reduces decision entropy.}
Fig.~\ref{fig:ent_fluc} (Right) shows the average entropy rate of token routing decisions across layers for our models compared to the baseline SMoE (on Wikitext-103), calculated at epoch 59, just before the final epoch where routing fluctuation occurs. Our Similarity/Attention-Aware SMoEs methods exhibit lower average entropy (rate < 1) than the baseline, consistent with the reduced routing fluctuation seen in the left graph. This indicates more stable and consistent routing decisions. The Similarity-Aware SMoE, in particular, maintains lower entropy across all layers, reflecting its better routing stability. These results highlight the advantages of our methods in enhancing routing consistency.

\textbf{Similarity/Attention-Aware SMoE improves load imbalancing.}
\begin{figure}[!t]
\centering
\includegraphics[width=0.75\linewidth]{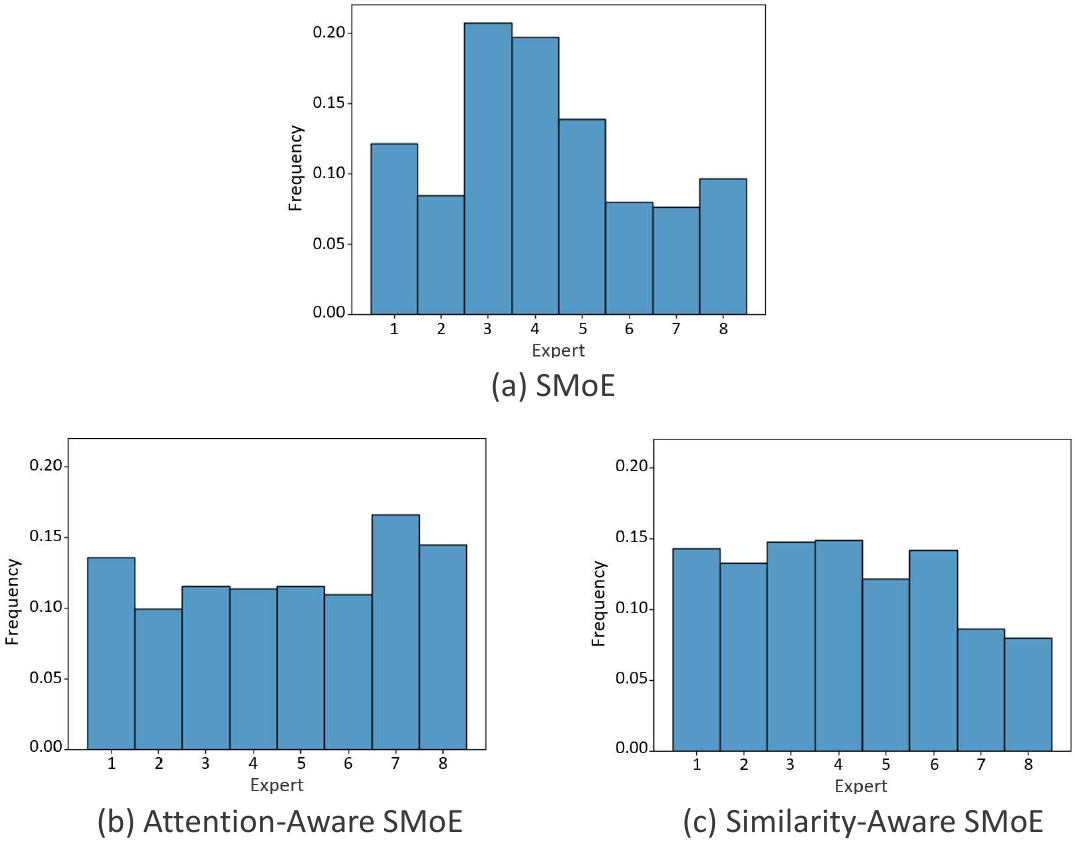}
\caption{\small Comparison of expert routing distribution for baseline SMoE, Attention-Aware SMoE, and Similarity-Aware SMoE}
\label{fig:load_im} 
\vspace{-.5cm}
\end{figure}
Fig.~\ref{fig:load_im} displays token distribution across experts in the VMoE architecture on the ImageNet validation set. The baseline model shows experts 3 and 4 handling significantly more tokens, while our Similarity/Attention-Aware SMoE models achieve a more uniform distribution. This implicit load balancing allows busier experts to specialize and ensures less utilized experts handle a broader range of inputs, improving overall efficiency and balance.
\section{Related Work}
\label{sec:related_work}
\textbf{Routing method.}
Numerous approaches have been proposed for assigning tokens to experts, including deterministic hashing~\cite{roller2021hash}, linear programming~\cite{lewis2021base}, and cosine similarity-based methods~\cite{chi2022representation}. Other techniques leverage reinforcement learning~\cite{bengio2015conditional}, greedy top-k expert selection~\cite{shazeer2017outrageously}, and optimal transport~\cite{liusparsity}. However, these approaches make expert assignment decisions for each token independently, without considering token-to-token interactions. In contrast, our proposed method enables collaborative expert selection by allowing tokens to share information with each other during the assignment process.

\textbf{Routing fluctuation.} Routing fluctuation as been discussed in existing literature.~\cite{nguyen2024libmoelibrarycomprehensivebenchmarking} mentions that various SMoE routers~\cite{csordas-etal-2023-approximating, do-etal-2023-hyperrouter} suffer from routing fluctuation without proposing solutions. In addition~\cite{su2024maskmoe} suggests that due to the variation of learnable parameters in the router. StableMoE~\citep{dai2022stablemoe} reduces routing fluctuations by first distilling a balanced routing strategy into a lightweight router and then locking token-to-expert assignments during the second training phase for stable routing.\textcolor{black}{SMoE-dropout~\cite{chen2023sparse} is another work that also provides another solution to improve the stability of the model. This method initially randomizes and freezes the router during training to provide stable routing strategies}  \cite{zoph2022st} examine several approaches to improve stability including removing multiplicative interactions, injecting model noise, and constraining activations and gradients. After the examination, the authors propose the router z-loss which enhance the training stability with no quality degradation. \cite{chi2022representation} proposes to estimate the routing scores between tokens and experts on a low-dimensional hypersphere to achieve more consistent routing compared to the conventional approach. Feedforward layers are replaced by hash layers in~\citep{roller2021hash} to to keep routing choices consistent. \cite{lewis2021base} formulates routing as
a linear assignment problem that globally maximizes token-expert similarities for increasing the stability. Our work is orthogonal to these approaches: to reduce routing fluctuation, we encourage tokens to influence each other's routing decision based on their similarity.  
\section{Concluding Remarks}
\label{sec:conclusion}
In this work, under the Probabilistic Graphical Model perspective, we show that in (S)MoE, expert selection are independent, results in routing fluctuations. To address this, we introduce Similarity/Attention-Aware (S)MoE, which establishes connections between tokens through a graph structure, effectively breaking this independence. Our theoretical analysis proves that both approaches reduce entropy in the decision-making process for indecisive tokens, leading to more stable and confident expert assignments. These theoretical advantages are validated through empirical evaluations across diverse tasks and domains, demonstrating the effectiveness of our proposed methods. A limitation of our paper is that we have not considered a generative model that capture the token generation process in our PGM. Studying transformer-MoE from a generative model perspective is an exciting research direction. We leave it for future work.

\section*{Impact Statement}
This paper presents work whose goal is to advance the field of 
Machine Learning. There are many potential societal consequences 
of our work, none which we feel must be specifically highlighted here.




\bibliography{example_paper}
\bibliographystyle{icml2025}

\newpage
\appendix
\onecolumn

\begin{center}
{\bf{\Large{Supplement to ``Improving Routing in\\ Sparse Mixture of Experts with Graph of Tokens"}}}
\end{center}

{\small
\startcontents
  \printcontents{ }{1}{\textbf{Table of Contents}\vskip3pt\hrule\vskip5pt}
  \vskip3pt\hrule\vskip5pt
}

\section{Notation}
\label{sec:notations}
To facilitate the understanding of the theoretical content of our work, all used notations are explained in \autoref{tbl:notations}.

\begin{table}[htbp]\caption{Table of notations}
\begin{center}
\begin{tabular}{r c p{10cm} }
\toprule
\multicolumn{3}{c}{\underline{Variables}}\\
$\bx_i$ & $\triangleq$ & input token of MHA at position $i$-th\\
$\bX$ & $\triangleq$ & $[\bx_1, \dots, \bx_N]^\top$, sequence of input token of MHA\\
$\bu_i$ & $\triangleq$ & input token of (S)MoE at position $i$-th\\
$\bU$ & $\triangleq$ &$[\bu_1, \dots, \bu_N]^\top$, sequence of input token of (S)MoE\\
$\bo_i$ & $\triangleq$ & target variable at position $i$-th\\
$h_i$ & $\triangleq$ & head selection at position $i$-th\\
$z_i$ & $\triangleq$ & attention position selection of token $\bx_i$\\
$s_i$ & $\triangleq$ & similarity variable of token $\bu_i$\\
$e_i$ & $\triangleq$ & expert selection of token $\bu_i$ for MoE generative model defined as in Section~\ref{sec:pgm_smoe}\\
$e^s_i$ & $\triangleq$ & expert selection of token $\bu_i$ for SAM defined as in Definition~\ref{def:sam}\\
$e^a_i$ & $\triangleq$ &expert selection of token $\bu_i$ for AAMM defined as in Definition~\ref{def:aamm_pgm}\\
\multicolumn{3}{c}{}\\
\multicolumn{3}{c}{\underline{Other Notations}}\\
\multicolumn{3}{c}{}\\
${N}$ & $\triangleq$ & number of tokens in a sequence\\
${H}$ & $\triangleq$ & number of heads in MHA\\
${E}$ & $\triangleq$ & number of experts in MoE\\
${K}$ & $\triangleq$ & number of selected experts in SMoE and its variants.\\
$\gamma_e(\bu_i)$ & $\triangleq$ & routing function giving the affinity score between expert $e$-th and $\bu_i$\\
$\rm{r}(\bu_i)$ & $\triangleq$ &$[r_1(\bu_i), \dots, r_N(\bu_i)]^\top = \mathrm{softmax}([\gamma_1(\bu_i), \dots, \gamma_K(\bu_i)]^\top)$ expert scores of token $\bu_i$ in (S)MoE\\
$\bm{r}_i$ & $\triangleq$ & $\bm{r}(\bu_i)$, short cut for $\bm{r}(\bu_i)$\\
$\bm{p}_i$ & $\triangleq$ & expert scores of token $\bu_i$ in Similarity/Attention-Aware (S)MoE\\
$\bm{g}_e$ & $\triangleq$ & expert network of expert $e$-th\\
$\mathbf{S}$ & $\triangleq$ & Similarity matrix of sequence $\bU$, whose elements are defined as in Definition~\ref{def:sam}\\
$\mathbf{A}_h$ & $\triangleq$ & Attention matrix of head $h$-th\\
$\mathbf{A}^p_h$ & $\triangleq$ & the posterior attention matrix whose element $\mathbf{A}^p_h[i,j] :=\bP(z_i = j | h_i = h, \bu_i, \bX)$ is defined as in Lemma~\ref{lem:finalda}\\
$\mathbf{H}^p$ & $\triangleq$ & poterior head selection matrx, whose element $\mathbf{H}^p[i, h] := \bP(z_i = j\mid h_i = h, \bX)$ represents the ``responsibility`` of head $h$ in explaining token $\bu_i$, defined as in Lemma~\ref{lem:finalda}\\
$\mathbf{L}_h$ & $\triangleq$ & the likelihood matrix, corresponding to head $h$, whose element $\mathbf{L}_{h}[i, j] := \mathcal{N}(\bu_i\mid\bW_h\bx_{j}, \sigma^2\mathbb{I})$ is defined as in Lemma~\ref{lem:finalda}.\\

\multicolumn{3}{c}{}\\
\multicolumn{3}{c}{\underline{Parameters}}\\
\multicolumn{3}{c}{}\\
$\bW, \bb$ & $\triangleq$ & weight matrix and bias vector to compute expert scores, respectively.\\
$\bW_{Q, h}, \bW_{K, h}, \bW_{V, h}, \bW_{O, h}$ & $\triangleq$ & query, key, value and output weight matrix used in MHA for head $h$-th, respectively\\
$\bW_{h}$ & $\triangleq$ & $\bW_{V, h}^\top\bW_{O,h}$ merging the value and output matrix for convenient notation\\
$\bW_{s}$ & $\triangleq$ & parameter matrix used to compute the similarity matrix $\mathbf{S}$ of sequence $\bU$\\
$\tau$ & $\triangleq$ & the temperature parameter used to compute the similarity between tokens of $\bU$\\
$\sigma$ & $\triangleq$ & the variance parameter of the distribution of $\bu_i$, given $e_i, h_i, \bX$, defined as in Definition~\ref{def:mam_pgm}.\\

\bottomrule
\end{tabular}
\end{center}
\label{tbl:notations}
\end{table}

\newpage
\section{Techincal Proofs}
\label{sec:proof}
\subsection{Proof of Lemma \ref{lem:finalda}}
{\color{black}
\label{sec:prooflem1}
\textbf{Restate Lemma~\ref{lem:finalda}}
    The expert decision of $e^a_i$ given $\bU, \bX$ is given by
    \begin{equation}
    \nonumber
        \bP(e^a_i = e | \bU, \bX) = \sum_{h = 1}^H \sum_{j = 1}^N \bold{H}^p[i, h]\bold{A}^p_{h}[i,j]r_e(\bu_{j})
    \end{equation}
where 
the posteriors
{
\begin{align*}
\nonumber
\bold{A}^{p}_{h}[i,j] &:= \bP(z_i = j | h_i = h, \bu_i, \bX) \\
&= \displaystyle \frac{\bold{A}_{h}[i, j]\mathbf{L}_{h}[i, j]}{\sum_{j'}\bold{A}_{h}[i, j']\mathbf{L}_{h}[i, j']},
\\
\bold{H}^{p}[i, h] &:=\bP(h_i = h | \bu_i, \bX) \\
&= \displaystyle\frac{\bold{H}[i, h] \sum_{j}\bold{A}_h[i, j]\mathbf{L}_h[i, j]}{\sum{h'}\bold{H}[i, h']\sum_{j'}\bold{A}_{h'}[i, j']\mathbf{L}_{h'}[i, j']},
\end{align*}}
with the prior $\bold{A}_{h}[i, j] := \bP(z_i = j| h_i = h, \bX)$ and $\bold{H}[i, h] := \bP(h_i = h)$ and the likelihood
$\mathbf{L}_{h}[i, j] := \mathcal{N}(\bu_i|\bW_h\bx_{j}, \sigma^2\mathbb{I})$. 
This results in
\begin{align}
\nonumber
    \mathbb{E}[\bo_i|\bX] = \mathbb{E}_{\bU}\left[\sum_{e = 1}^E\bP(e^a_i = e | \bU, \bX)\mathbf{g}_{e}(\bu_i)| \bX\right]
\end{align}

\textbf{Review the generalization of the graph $\mathcal{G}_4$}. 
Given the sequence of input $\bX = [\bx_1, \dots, \bx_N]^{\top}$, let $h_i \in \{1, \dots, H \}$ and $z_i \in \{1, \dots, N\}$ are the selected head and attention position of each token $i$, respectively. Let $U = [\bu_1, \dots, \bu_N]^N$ be the sequence of latent variable and $e^a_i \in \{1, \dots, E\}$ is the expert assignment for token $i$. AAMM generates the output $\bo_i$ as follow
    \begin{equation}
    \begin{aligned}
    \nonumber
    \label{eq:moe_mha_pgm}
h_i &\sim \mathrm{Uniform}(\{1,\ldots,H\}) \qquad \\
z_i|h_i,\mathbf{X} &\sim \mathrm{Cat}\left(\mathrm{Softmax}\Bigl(\frac{\bx_i^{\top}\bW_{Q, h_i}^{\top}\bW_{K, h_i}\bX^T}{\sqrt{D}} \Bigl)\right) \qquad \\
\bu_i|{z}_i, h_i,\bX &\sim \mathcal{N}(\bW_{h_i}\bx_{z_i}, \sigma^2\mathbb{I})\\
e^a_i|z_i,\bU &\sim \mathrm{Cat}(\bm{r}(\bu_{z_i})) \\
\bo_i|\bu_i,e^a_i &\sim \mathcal{N}(\mathbf{g}_{e^a_i}(\bu_i), \mathbb{I}) \qquad
    \end{aligned}
\end{equation}

\textbf{Proof}: Following the above generative process, starting with the probability of expert decision $e^a_i$ of AAMM  for token $i$ given the sequence $\bU$ and $\bX$:
{\color{black}{\begin{equation}
\label{eq: lem1_proof}
    \begin{aligned}
        \bP(e^a_i = e | \bU, \bX) &= \sum_{j = 1}^N\bP(e^a_i = e | z_i = j, \bU, \bX)\bP(z_i = j | \bU, \bX)\\
        &= \sum_{j = 1}^N\bP(e^a_i = e | z_i = j, \bU)\sum_{h = 1}^H\bP(z_i = j, h_i = h | \bU, \bX )\\
        &= \sum_{j = 1}^N\bP(e^a_i = e | z_i = j, \bU)\sum_{h = 1}^H\bP(z_i = j | h_i = h, \bU, \bX)\bP(h_i| \bU, \bX)\\
        &= \sum_{j = 1}^N\bP(e^a_i = e | z_i = j, \bU)\sum_{h = 1}^H\bP(z_i = j | h_i = h, \bu_i, \bX)\bP(h_i| \bu_i, \bX)
    \end{aligned}
\end{equation}}}


The posterior distribution of attention variable given the observation of MoE input $\bu_i$ is 
{\color{black}{\begin{equation}
    \begin{aligned}
    \bP(z_i = j| h_i = h, \bu_i, \bX) &=  \frac{\bP(z_i = j| h_i = h, \bX)\bP(\bu_i| z_i = j, h_i = h, \bx_j)}{\sum_{j'}\bP(z_i = j'| h_i = h, \bX)\bP(\bu_i| z_i = j', h_i = h, \bx_j')}\\
        &=\displaystyle \frac{\bold{A}_h[i, j]\mathcal{N}(\bu_i|\bold{W}_{O,h}\bold{W}_{V,h}\bx_{j}, \mathbb{I})}{\sum_{j'}\bold{A}_h[i, j']\mathcal{N}(\bu_i|\bold{W}_{O,h}\bold{W}_{V,h}\bx_{j'}, \mathbb{I})} \\
        &= \displaystyle \frac{\bold{A}_h[i, j]\mathbf{L}_h[i, j]}{\sum_{j'}\bold{A}_h[i, j']\mathbf{L}_h[i, j']} = \bold{A}^{p}_h[i,j]
    \end{aligned}
\end{equation}}}
where $\bA_h$ is the attention matrix of head $h$ and the likelihood $\mathbf{L}_h[i, j] = \mathcal{N}(\bu_i|\bW_h \bx_{j}, \mathbb{I})$
\\
Then, the posterior probability of head index given input $\bu_i$ of the (S)MoE and input $\bX$ of the attention. This represents the responsibility of head $h$ in explaining token $i$. 
{\color{black}{\begin{equation}
\begin{aligned}
    \bP(h_i = h | \bu_i, \bX) &= \frac{\bP(h_i = h) \sum_{j = 1}^N \bP(z_i = j| h_i = h, \bX)\bP(\bu_i| z_i = j, h_i = h,\bx_j)}{\sum_{h' = 1}^H\bP(h_i = h') \sum_{j' = 1}^N \bP(z_i = j'| h_i = h', \bX)\bP(\bu_i| z_i = j', h_i = h',\bx_{j'})}\\
    &=\displaystyle\frac{\bold{H}[i, h] \sum_{j}\bold{A}_h[i, j]\mathcal{N}(\bu_i|\bold{W}_{O,h'}\bold{W}_{V,h'}\bx_{j}, \mathbb{I})}{\sum{h'}\bold{H}[i, h']\sum_{j'}\bold{A}_{h'}[i, j']\mathcal{N}(\bu_i|\bold{W}_{O,h'}\bold{W}_{V,h'}\bx_{j'}, \mathbb{I})}\\
    &=\displaystyle\frac{\bold{H}[i, h] \sum_{j}\bold{A}_h[i, j]\mathbf{L}_{h}[i, j]}{\sum{h'}\bold{H}[i, h']\sum_{j'}\bold{A}_{h'}[i, j']\mathbf{L}_{h'}[i, j'], \mathbb{I})}
    = \bold{H}^{p}[i, h].
\end{aligned}
\end{equation}}}
Substituting the results to Eqn.~\ref{eq: lem1_proof}, we obtain:
 \begin{equation}
    \nonumber
        \bP(e^a_i = e | \bU, \bX) = \sum_{h = 1}^H \sum_{j = 1}^N \bold{H}^p[i, h]\bold{A}^p_{h}[i,j]r_e(\bu_{j})
    \end{equation}
To compute $\mathbb{E}(\bo_i| \bX)$, we condition $\bo_i$ on $e^a_i$, $\bU$, and $\bX$, results in:
\begin{equation}
    \begin{aligned}
    \nonumber
    \mathbb{E}[\bo_i|\bX] &= \mathbb{E}_{\bU}\left[\mathbb{E}_{e^a_i}\left[\mathbb{E}_{\bo_i}[\bo_i| e^a_i,\bU] | \bU, \bX \right] | \bX \right]\\
&=\mathbb{E}_{\bU}\left[\mathbb{E}_{e^a_i}\left[\mathbb{E}_{\bo_i}[\bo_i| e^a_i,\bu_i] | \bU, \bX \right] | \bX \right]\\
&=\mathbb{E}_{\bU}\left[\sum_{e} \bP(e^a_i = e | \bU, \bX) \mathbf{g}_{e}(\bu_i) | \bX \right]\\
    \end{aligned}
\end{equation}

Thus, we have derived the complete dependency of $e^a_i$ on $\bU$ and $\bX$ through the expert selection process of Attention-Aware SMoE , proving Lemma 1.
}
\subsection{Proof of Proposition \ref{prop:entroy}}
{\color{black}{
\textbf{Restate Proposition~\ref{prop:entroy}}
\label{sec:proofprop1}
\setcounter{prop}{0}
\begin{prop}
\label{prop:entroy_supp}
Let $\bm{p}_i = [p_1, \dots, p_K]^T$ denote the distribution of expert selection variables, i.e., $e^s_i$ for SAM and $e^a_i$ for A$^2$MM. The expert score in the baseline MoE for token $\bu_i$ is breviated as $\bm{r}_i := \bm{r}(\bu_i)$ as in Section~\ref{subsec:SMoE}. 
Similarity/Attention-Aware MoE transform these expert  scores $\bm{r}_i$ into $\bm{p}_i = \sum_{j \in J_i} s(i,j)\bm{r}_j$, where $s(i,j)$ denotes the influence weight between tokens $\bu_i$ and $\bu_j$. 
The upper bound of the expert selection's entropy in Similarity/Attention-Aware MoE is then given by:
    \begin{equation}
        \mathcal{H}(\bm{p}_i) \leq \sum_{j = 1}^{|J_i|}s(i, j)\mathcal{H}(\bm{r}_j) + \mathcal{H}(\bold{s}_i),
    \end{equation}
    where $\bold{s}_i = [s(i, 1), \dots, s(i, |J_i|)]^T$. As the temperature parameter $\tau \to 0$ (defined in Def.~\ref{def:sam}) or $\sigma \to 0$ (defined in Def.~\ref{def:aamm_pgm}), $\mathcal{H}(\bm{p}_i) \leq \mathcal{H}(\bm{r}_i)$. 
\end{prop}
}}
\textbf{Proof:} We first prove for the case of Similarity-Aware routing. The proof is the same for Attention-Aware routing.
From $\bm{p}_i =\displaystyle \sum_{j = 1}^{|J_i|}s(i, j){\bm{r}}_j$, we have $e^s_i$ is the mixture of $|J_i|$ discrete distribution of $e_j$ with the probability mass $\bm{r}_i$. Denote $t_i$ is the latent random variable of that admit the weighting coefficient as probability distribution $s(i, .)$ at token $i$. We obtain the decomposition of joint entropy as follow
\begin{align}
    \nonumber 
    \mathcal{H}(e^s_i, t_i) = \mathcal{H}(t_i) + \mathcal{H}(e^s_i | t_i) = \mathcal{H}(t_i) + \sum_{j = 1}^{|J_i|}s(i, j)\mathcal{H}(e_j)
\end{align}
{\color{black} Since entropy is non-negative,}
\[ \mathcal{H}(e^s_i, t_i) = \mathcal{H}(e^s_i) + \mathcal{H}(t_i| e^s_i) \geq \mathcal{H}(e^s_i)\]
Hence, 
\[
\mathcal{H}(e^s_i) \leq \mathcal{H}(t_i) + \sum_{j = 1}^{|J_i|}s(i, j)\mathcal{H}(e_j) \leq \mathcal{H}(t_i) + \mathcal{H}(e_i)
\]
because for any $j \in J_i$, $\mathcal{H}(e_i) > \mathcal{H}(e_j)$.
 {\color{black} Again, here, we slightly abuse the notation of entropy $\mathcal{H}$, using it interchangeably for both a random variable and its associated distribution.}

The final piece of this Proposition's proof is to verify the above limit. For $\tau\to 0$, the temperature-softmax distribution gradually morphs into an one-hot distribution, and thus $\mathcal{H}(t_i)$ entropy goes to 0. Therefore, when $\tau \to 0$, we have $\mathcal{H}({e^s_i}) \leq \mathcal{H}(e_i)$ or $\mathcal{H}(\bm{p}_i) \leq \mathcal{H}({\bm{r}}_i)$. We have proved the inequality in Prop.~\ref{prop:entroy} for Similarity-Aware MoE

Similarly, for Attention-Aware routing , we also show that $\sigma \to 0$, we have $\mathcal{H}({e^a_i}) \leq \mathcal{H}(e_i)$ or $\mathcal{H}(\bm{p}_i) \leq \mathcal{H}({\bm{r}}_i)$. As $\sigma\to 0$, $\mathbb{P}(\bu_i | z_i = j, h_i = h, \bX) = \mathcal{N}(\bu_i| \bW_h\bx_j, \sigma^2\mathbb{I})$ also converges to the Dirac delta function centered at the mean. This means that the closest mean will give a density greatly dominating the others, in turn making $\bm{A}_{h^*}^p$ the one-hot distribution, yielding zero entropy of $\mathcal{H}(t_i| e^s_i)$.

With that, we have fully proved Prop~\ref{prop:entroy_supp}.

\subsection{ the equivalence of $\mathrm{Renormalize}(\mathrm{TopK^0}
(r_e))$ and $\mathrm{Softmax}(\mathrm{TopK}^{\infty}(\gamma_e))$, and its proof}
\label{subsec:renorm}
{\color{black} \textbf{Renormalization.} 
Given a token $\bu_i$, let $\mathcal{K}$ be the set of indices with $K$ highest expert scores $r_k(\bu_i)$. It is also equivalent to the set of $K$ highest affinity score $\gamma_k(\bu_i)$.
For any $e \in \mathcal{K}$, we have
\begin{equation}
\nonumber
    \begin{aligned}
        \mathrm{Renormalize}(\mathrm{TopK^0}(r_e)) &= \frac{r_e(\bu_i)}{\sum_{k \in \mathcal{K}}r_k(\bu_i)} \\
        &= \displaystyle \frac{\mathrm{exp}^{ \gamma_e(\bu_i)}/\sum_{e' = 1}^E \mathrm{exp}^{\gamma_e'(\bu_i)}}{\sum_{k \in \mathcal{K}}\mathrm{exp}^{\gamma_k(\bu_i)}/\sum_{e' = 1}^E \mathrm{exp}^{\gamma_e'(\bu_i)}}\\
        &= \displaystyle \frac{\mathrm{exp}^{ \gamma_e(\bu_i)}}{\sum_{k \in \mathcal{K}}\mathrm{exp}^{\gamma_k(\bu_i)}} = \mathrm{Softmax}(\mathrm{TopK}(\gamma_e(\bu_i)))\\
    \end{aligned}
\end{equation}
This shows the equivalence of two operators for $e \in \mathcal{K}$
For any $e \notin \mathcal{K}$, $\mathrm{TopK^0}(r_e(\bu_i)) = \mathrm{exp}(\mathrm{TopK}^{\infty}(\gamma_e(\bu_i))) = 0$ also results in the equivalence of two operators, finishing the proof.

\section{Derivation}
\label{sec:derivation}
\textbf{Conditional Expectation of $\bu_i$ given $\bX$ following MAM}

By using the tower rule, we obtain
\begin{equation}
\begin{aligned}
\mathbb{E}[\bu_i|\bX] &= \mathbb{E}_{h_i}\bigl[ \mathbb{E}_{z_i}\big[ \mathbb{E}_{\bu_i}[\bu_i| z_i, h_i, \bX]| h_i, \bX\big]\bigl]\\
&= \sum_h^H \sum_j^N \mathbb{E}[\bu_i| z_i = j, h_i, \bx_j] \bP(z_i = j| h_i = h, \bX)\bP(h_i = h) \nonumber\\
&=\frac{1}{H} \sum_{h=1}^{H} \mathbf{W}_{h} \sum_{j=1}^{N} \text{Softmax}\left( \frac{\bx_i^{\top}\bW_{Q, h_i}^{\top}\bW_{K, h_i}\bx_j}{\sqrt{D_{qk}}} \right) \mathbf{x}_{j}.    
\end{aligned}
\end{equation}
\section{Experiments Details}
\begin{table}
\centering
\caption{Summarization of baselines' sizes}
\label{tab:baseline-sizes}
\vspace{0.5em}
\scalebox{.9}{
\begin{tabular}{lcccccc}
\hline
Baselines size & \# params & \# layers & Token dimension $D$ & Hidden size of MLP $d_\text{ff}$ & Sequence Length & \# heads \\
\hline
VMoE & 60M & 8 & 512 & 2048 & 50 & 8 \\
SMoE-medium & 215M & 6 & 352 & 352 & 512 & 8 \\
SMoE-large & 388M & 12 & 512 & 512 & 512 & 8 \\
GLAM-medium & 201M & 4 & 352 & 351 & 512 & 8 \\
\hline
\end{tabular}}
\end{table}
\label{sec:expdetails}
\subsection{WikiText-103 Language Modeling}
\subsubsection{Dataset} The WikiText-103 dataset \cite{merity2016pointersentinelmixturemodels}, sourced from Wikipedia, is crafted to examine extended contextual relationships. Its training component encompasses roughly 28,000 articles, totaling 103 million words. These articles are segmented into blocks of about 3,600 words each. The validation and test sets consist of 60 articles each, with word counts of 218,000 and 246,000 respectively, amounting to approximately 268,000 words combined.
To assess the resilience of our methods, we employ TextAttack's word swap attack \cite{morris2020textattack} to modify both the validation and test data. This adversarial method randomly substitutes words with "AAA," challenging the model's ability to accurately predict subsequent words in the sequence.
\subsubsection{Models and baselines}
In our study, we utilize the Switch Transformer \cite{fedus2021switch} (denoted as SMoE in our data presentations) and GLaM \cite{du2022glam} as baseline models. The Switch Transformer substitutes all multilayer perceptron (MLP) layers with SMoE layers, while GLaM replaces every alternate MLP layer.
Our standard model for experiments is medium-sized with 6 layers.
Each model incorporates 16 experts in every models, selecting Top-1 or Top-2 experts (E = 2) per input. All models employ an identical sparse router function, comprising a linear network that processes input data, followed by TopK and Softmax functions.
The  models undergo 60 epochs of training, while GLaM models train for 80 epochs without any additional load balancing loss.
Our implementation builds upon the codebase developed by \cite{pham2024competesmoe,press-etal-2020-improving}, which is publicly accessible at \href{https://github.com/ofirpress/sandwich\_transformer}{https://github.com/ofirpress/sandwich\_transformer} and \href{https://github.com/giangdip2410/CompeteSMoE/tree/main}{https://github.com/giangdip2410/CompeteSMoE/tree/main}.

The model sizes are summarized in Tab.~\ref{tab:baseline-sizes}. Note that except for models presented in Tab.~\ref{tab:model_size_appendix}, all models used in language tasks have medium sizes. 

In all our Similarity/Attention-Aware SMoEs, we set the hyperparameter $\tau = 1$. In Similarity-Aware SMoE, instead of learning $\bW_s$ in Def.~\ref{def:sam}, we set $\bW_s = \mathbb{I}$ for the save of computation and to avoid introduce extra parameters. In Attention-Aware SMoE , we set the hyperparameter $\sigma=1$.
\subsection{ImageNet-1K Object Recognition}
\label{sec:k=1}
\subsubsection{Datasets}
Our study employs the ImageNet-1K dataset, which consists of 1.28 million training images and 50,000 validation images across 1,000 object classes. The model is trained for object recognition.
To evaluate resilience to input data distribution shifts, we use ImageNet-A (IN-A) \cite{hendrycks2021natural}. This dataset includes adversarially filtered images from a 200-class subset of ImageNet-1K.
We also test our model's ability to generalize to abstract visual representations using ImageNet-R (IN-R) , which contains various artistic renditions of images.
\subsubsection{Model and baselines}
For our ImageNet-1K object recognition task and standard robustness benchmarks, we employ a small Vision Mixture of Experts (V-MoE) \cite{riquelme2021scaling} model as the SMoE baseline. This V-MoE variant is composed of 8 Vision Transformer (ViT) blocks, with the MLPs in the final two blocks replaced by SMoE layers.
In our Similarity/Attention-Aware SMoEs, we alternate between Attention-Aware SMoE  and Similarity-Aware SMoE layers, replacing every other MLP layer. All our vision SMoE models select 2 experts ($K = 2$) per patch at each SMoE layer.
We adhere to the training configurations and settings outlined in the cited work. The codebase for this implementation is publicly available at \href{https://github.com/google-research/vmoe/}{https://github.com/google-research/vmoe/}. Similar to the experiments on Language Modeling, we also we set the hyperparameter $\tau = 1$ and $\bW_s = \mathbb{I}$ in Similarity-Aware SMoE.

\textcolor{black}{The VMoE baseline has 8 layers, with model size is 512 and 60M parameters.}
\subsection{Finetuning on downstream tasks.}
\subsubsection{Datasets}

\textbf{The Stanford Sentiment Treebank-2 (SST2).}~\cite{socher-etal-2013-recursive}. The dataset is designed for analyzing how sentiment is composed in language. It contains a binary classification setup, featuring 11,855 individual sentences drawn from movie reviews. Using the Stanford parser, these sentences were broken down into parse trees, generating 215,154 distinct phrases. Each of these phrases was evaluated by three human annotators. The dataset's structure allows researchers to examine how sentiment meaning is built up through language composition.

\textbf{The Stanford Sentiment Treebank-5 (SST5)}~\cite{socher-etal-2013-recursive}. The sentiment analysis dataset consists of five sentiment categories. It comprises 11,855 individual sentences taken from movie reviews. The sentences were parsed into trees, yielding 215,154 unique phrases, with each phrase receiving ratings from three human evaluators. The sentiment classifications in this dataset range across five levels: negative, somewhat negative, neutral, somewhat positive, and positive.

\textbf{The Banking-77 (B77)}~\cite{Casanueva2020} This is a detailed intent classification dataset for customer service in banking. It features 13,083 customer queries categorized across 77 distinct intent classes, providing a highly granular classification system for banking-related customer inquiries.

\subsubsection{Model and baselines}
The models were initialized using pretrained language models that were trained on the Wikitext-103 dataset. For the Stanford Sentiment Treebank datasets (SST2 and SST5), the model training process involves 5 epochs of finetuning, utilizing the Adam optimizer with a 0.001 base learning rate. The training uses no warmup period and processes 16 samples per batch. The Banking-77 dataset requires longer training at 50 epochs, also using Adam but with a much lower base learning rate of 0.00001, maintaining the same batch size of 16 and no warmup period.

\section{Additional Experiments and Analysis}
\label{sec:add_exp}
\subsection{Experiments with change in number of experts and $\mathrm{TopK}$}
We examine our method with more seetings of number of experts and $\mathrm{TopK}$ including Top-1, Top-8 and 32 experts. Across all these settings, both Similarity-Aware SMoE and Attention-Aware SMoE consistently demonstrate better performance compared to the baseline SMoE, achieving lower PPL scores on both Clean and Attacked Wikitext-103 datasets (Tab ~\ref{tab:expert_appendix}). When using 32 experts, our methods achieve PPL reductions of up to 1.56 PPL compared to the baseline, and when increasing to top-8 active experts, they maintain their advantage with improvements of up to 1.64 PPL. These consistent performance gains across different architectural configurations demonstrate the robustness and effectiveness of our proposed methods.
\begin{table}[!t]
\centering

\caption{\small \color{black} PPL evaluation (lower is better) with the clean and attacked Wikitext-103 test set of baseline SMoEs and Similarity/Attention-Aware-SMoE SMoE(s) with different number of experts and $\mathrm{TopK}$}

    \vspace{-0.5em}
    \color{black}
    \begin{center}
    \scalebox{.95}{
    \begin{tabularx}{\linewidth}{lYYYY}
    \toprule
         \multirow{2}{*}{Model/Metric} & \multicolumn{2}{c}{Clean Wikitext-103} & \multicolumn{2}{c}{Attacked Wikitext-103} \\
         \cmidrule(r){2-3}\cmidrule(r){4-5}
          & Valid PPL & Test PPL & Valid PPL & Test PPL\\
        \midrule
        
       \textcolor{black}{\it SMoE (K = 1, E = 16)} & \textcolor{black}{39.55}  & \textcolor{black}{40.75} & \textcolor{black}{48.82}  & \textcolor{black}{50.21}\\
\textcolor{black}{Similarity-Aware SMoE (K = 1, E = 16)} &  \textcolor{black}{37.78} & \textcolor{black}{39.18} &  \textcolor{black}{46.93} & \textcolor{black}{48.66}\\
\textcolor{black}{Attention-Aware SMoE  (K = 1, E = 16)} & \textcolor{black}{38.02} & \textcolor{black}{39.35} & \textcolor{black}{47.20} & \textcolor{black}{48.72}\\
\midrule
\color{black}
\it \textcolor{black}{SMoE (K = 2, E = 16)} & \textcolor{black}{33.29}  & \textcolor{black}{34.84}  & \textcolor{black}{41.75}  & \textcolor{black}{43.59} \\
\textcolor{black}{Similarity-Aware SMoE (K = 2, E = 16)} & \bf \textcolor{black}{30.75} & \bf \textcolor{black}{32.03} &  \bf \textcolor{black}{38.33} & \bf \textcolor{black}{39.92}\\
\textcolor{black}{Attention-Aware SMoE  (K = 2, E = 16)} & \textcolor{black}{31.31}  & \textcolor{black}{32.23} &  \textcolor{black}{39.68} & \textcolor{black}{40.91}\\
        \midrule
       \it SMoE (K = 8, E = 16) & 33.48&	34.92&	41.36&	42.98\\
        Similarity-Aware SMoE (K = 8, E = 16) & 32.5&	33.81&	40.6&	42.37\\
        Attention-Aware SMoE  (K = 8, E = 16) &\textbf{31.97}	&\textbf{33.28}&	\textbf{39.98}&	\textbf{41.45}\\
        \midrule
        \it SMoE (K = 2, E = 32) & 31.82&	33.41&	39.9&	41.79\\
        Similarity-Aware SMoE (K = 2,, E = 32) & 30.41&	\textbf{31.62}&	38.23	&39.77\\
        Attention-Aware SMoE  (K = 2, E = 32) &\textbf{30.39}&	31.85&	\textbf{37.8}&	\textbf{39.65}\\
        \bottomrule
    \end{tabularx}}
    \end{center}
\vspace{-0.15in}

\label{tab:expert_appendix}
\end{table}
\subsection{Routing Fluctuation and Entropy of SMoEs Top-1}
\label{subsec:routing-top1}
\begin{figure}[!t]
\vspace{5mm}
\centering
\includegraphics[width=0.9\textwidth]{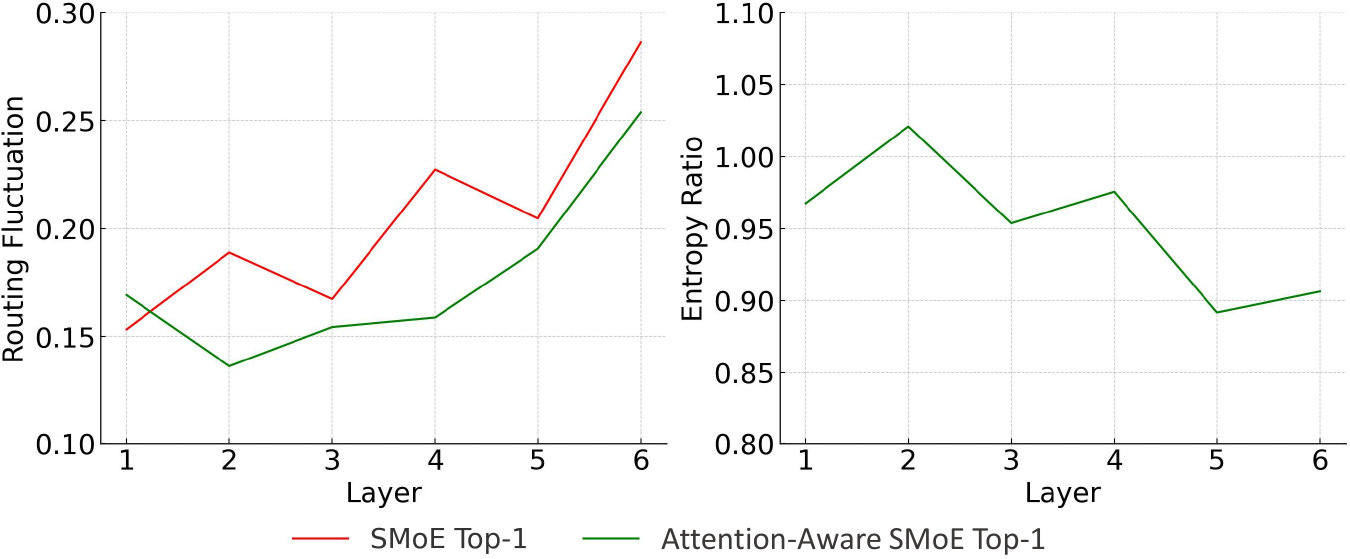}
\vspace{-3mm}
\caption{\small Comparison of Routing Fluctuation and Entropy Ratio Across Layers for Baseline SmoE Top-1, Attention-Aware SMoE  Top-1, and Similarity-Aware SMoE Top-1}
\label{fig:ent_fluc_top1} 
\end{figure}
\textbf{Attention-Aware SMoE Top-1 reduces routing fluctuation.}
Fig. \ref{fig:ent_fluc_top1} (Left) compares the routing fluctuation of the baseline SMoE Top-1 and Attention-Aware SMoE  for Top-1 routing. The fluctuation rate, computed as the proportion of tokens that switch their expert choice between consecutive last training epochs (from epoch 59 to 60), provides insight into routing stability. The Baseline SMoE exhibits higher fluctuation rates across all layers. In contrast, the Attention-Aware SMoE  demonstrates consistently lower fluctuation rates across all layers. The Attention-Aware SMoE  maintains more stable routing decisions throughout the network, indicating improved consistency in expert utilization. These results suggest that our proposed Attention-Aware SMoE method significantly enhances routing stability compared to the baseline approach, potentially leading to more consistent and efficient utilization of experts in the Mixture of Experts model. The results also aligns with the better performance and enhancement in robustness of Attention-Aware SMoE  Top-1 in Tab. \ref{tab:expert_appendix}.

\textbf{Attention-Aware SMoE Top-1 reduces decision entropy}
Fig. \ref{fig:ent_fluc_top1} (Right) illustrates the ratio of average entropy of tokens' routing decisions across layers for the Attention-Aware SMoE  compared to the baseline SMoE for epoch 59. The Attention-Aware SMoE  demonstrates consistently lower entropy levels compared to the baseline SMoE across all layers, as evidenced by ratios below 1.0. This trend aligns with the lower routing fluctuation observed in the left graph, suggesting that our approach leads to more stable and consistent routing decisions.

\subsection{Scalability of Similarity/Attention-Aware SMoEs}
\label{subsec:scale}
\begin{table}[t!]
\centering
\caption{\small \color{black} PPL evaluation (lower is better) with the clean and attacked Wikitext-103 test set Baseline SMoE (large size), Attention-Aware SMoE  (large size),  and Similarity-Aware SMoE (large size)}

    \vspace{-0.5em}
    \color{black}
    \begin{center}
    \scalebox{.95}{
    \begin{tabularx}{\linewidth}{lYYYY}
    \toprule
         \multirow{2}{*}{Model/Metric} & \multicolumn{2}{c}{Clean Wikitext-103} & \multicolumn{2}{c}{Attacked Wikitext-103} \\
         \cmidrule(r){2-3}\cmidrule(r){4-5}
          & Valid PPL & Test PPL & Valid PPL & Test PPL\\
        \midrule

        \it SMoE (K = 2) & 28.737&	30.378&	36.43&	38.34 \\
        Similarity-Aware SMoE (K = 2) & \textbf{27.06}&	\textbf{28.34}&	\textbf{34.65}&	\textbf{36.28}\\
        Attention-Aware SMoE  (K = 2) & 27.26&	28.69	&34.69&	36.37\\
        \bottomrule
    \end{tabularx}}
    \end{center}
\vspace{-0.15in}

\label{tab:model_size_appendix}
\end{table}
To further demonstrate the scalability of our models, we evaluate them with a larger transformer-MoEs baseline of approximately 390M parameters, with 12 layers. The results in Tab.~\ref{tab:model_size_appendix} confirms that the scaling law holds true, as all models show improved language modeling performance with increased parameter count. Importantly, both Similarity-Aware SMoE and Attention-Aware SMoE maintain their performance advantage over the conventional SMoE at this larger scale, with Similarity-Aware SMoE emerging as the best performing variant. These findings validate that the benefits of our proposed methods are preserved when scaling up model size.

\subsection{Comuputation and memory}
\label{subsec:computation}
\begin{table}[t!]

\caption{\small \color{black} Computation and Memory Ratio of forward pass (compared to the baselines SMoE, XMoE and SMoE-dropout) comparison for different SMoE-medium size variants, Top-K = 2}
    \vspace{-0.5em}
    \color{black}
    \begin{center}
    \scalebox{.95}{
    \begin{tabularx}{\linewidth}{lYY}
    \toprule
         Model & Computation Ratio & Memory Ratio \\
        \midrule
        Similarity-Aware SMoE & 1.048 & 1.008 \\
        Attention-Aware SMoE  & 1.070 & 1.060 \\
        \midrule
        Similarity-Aware SMoE XMoE & 1.026 & 1.009 \\
        Attention-Aware SMoE XMoE & 1.038 & 1.060 \\
        \midrule
        Similarity-Aware SMoE-dropout & 1.047 & 1.008 \\
        Attention-Aware SMoE -dropout & 1.064 & 1.060 \\
        \bottomrule
    \end{tabularx}}
    \end{center}
\label{tab:moe_ratios}
\end{table}
{\color{black}{
 We compare the computational complexity and memory complexity of using mutual inform techniques compared to the conventional approach without them. In particular, we measure the computational time and computational memory of  Similarity-Aware SMoE and Attention-Aware SMoE divided by the corresponding computational time and computational memory of the conventional SMoE in Tab.~\ref{tab:moe_ratios}. Similarly, we report the ratio for the case of XMoE and SMoE-dropout in Tab.~\ref{tab:moe_ratios}. From the table, we can see that Similarity/Attention-Aware-SMoE variants only increase the computational complexities slightly. 
}
}

\subsection{Hyperparameter ablation}
\label{subsec:hyper}
We present the ablation study for the hyperparameters temperatures $\tau$ in Similarity-SMoE and $\sigma$ in Attention-SMoE. Tab.\ref{tab:hyper} demonstrates that both Similarity-SMoE and Attention-SMoE are relatively insensitive to their respective temperature parameters ($\tau$ and $\sigma$). Across different values including $0.1, 1, 2$, and $\sqrt{352}$ (where 352 is the model size). In the case of Similarity-SMoE, too large $\tau$ or too small $\tau$ can lead to an decrease in performance.
\begin{table}[t!]
\caption{\small \color{black} Perplexity comparison for different SMoE variants with various $\tau$ values on validation and test sets}
    \color{black}
    \begin{center}
    \scalebox{.85}{
    \begin{tabularx}{\linewidth}{lYY}
    \toprule
         Model & Valid PPL & Test PPL \\
        \midrule
        SMoE & 33.29 & 34.84 \\
        \midrule
        Similarity-SMoE ($\tau$=0.1) & 32.79 & 34.01 \\
        Similarity-SMoE ($\tau$=1.0) & 30.75 & 32.03 \\
        Similarity-SMoE ($\tau$=2.0) & 30.68 & 32.88 \\
        Similarity-SMoE ($\tau$=$\sqrt{352}$) & 32.26 & 33.83 \\
        \midrule
        Attention-SMoE ($\sigma$=0.1) & 31.93 & 32.67 \\
        Attention-SMoE ($\sigma$=1.0) & 31.31 & 32.23 \\
        Attention-SMoE ($\sigma$=2.0) & 31.13 & 32.85 \\
        Attention-SMoE ($\sigma$=$\sqrt{352}$) & 31.62 & 32.90 \\
        \bottomrule
    \end{tabularx}}
    \end{center}
\label{tab:hyper}
\end{table}
\subsection{The influence of token similarity on the expert decision during training}
\label{subsec:dynmaic_tau}
The token similarity influences decisions through the score matrix $S = \mathrm{Softmax}(\bU\bW_s\bU^T/\tau)$, where $\tau$ is the temperature parameter and in practice, we set $\bW_s = \mathbb{I}$. As $\tau \to 0$, $S$ converges to the identity matrix, meaning each token becomes conditionally independent in decision-making. We examine how gradually varying $\tau$ during training affects model performance. As expected, when $\tau \to 0$, the performance of Similarity-Aware SMoE becomes comparable with the baseline SMoE. On the other hand, increasing $\tau$ from 0.01 to 10 improves the performance to the baseline.
\begin{table}
\centering
\caption{PPL evaluation (lower is better) with the clean and attacked Wikitext-103 test set of medium-size baseline SMoEs and Similarity-Aware SMoE(s) with $\tau$ decreases from 10 to 0.01 and increases from 0.01 to 10}
\label{tab:ppl-evaluation}
\scalebox{.8}{
\begin{tabularx}{\linewidth}{lYYYY}
    \toprule
         \multirow{2}{*}{Model/Metric} & \multicolumn{2}{c}{Clean Wikitext-103} & \multicolumn{2}{c}{Attacked Wikitext-103} \\
         \cmidrule(r){2-3}\cmidrule(r){4-5}
          & Valid PPL & Test PPL & Valid PPL & Test PPL\\
        \midrule
SMoE & 33.29 & 34.84 & 41.75 & 43.59 \\
Similarity-Aware SMoE ($\tau$ decreases) & 33.37 & 34.65 & 41.61 & 43.08 \\
Similarity-Aware SMoE($\tau$ increases) & 32.40 & 33.91 & 39.92 & 41.67 \\
\bottomrule
\end{tabularx}}
\end{table}





\end{document}